\documentclass[journal]{IEEEtran}
\usepackage{}
\usepackage{lineno,hyperref}
\usepackage{graphicx}
\usepackage{epstopdf}
\usepackage{mathrsfs}
\usepackage{amsfonts}
\usepackage{bm}
\usepackage{multirow}
\usepackage{threeparttable}
\usepackage[subfigure]{tocloft}
\usepackage{subfigure}
\usepackage{amsmath}
\usepackage{afterpage}
\usepackage{float}
\usepackage[linesnumbered,ruled,vlined]{algorithm2e}
\usepackage{placeins} 
\usepackage{booktabs}
\usepackage[table,xcdraw,dvipsnames]{xcolor}

\usepackage{booktabs}   
\usepackage{xcolor}     
\usepackage{graphicx}   

\definecolor{myPink}{rgb}{0.9294, 0.0078, 0.5490}
\definecolor{Gray}{gray}{0.92}
\definecolor{my_color}{HTML}{E8F3F1}
\definecolor{my_color1}{HTML}{FFEACE}
\definecolor{my_color2}{HTML}{FBEAFF}
\definecolor{my_color3}{HTML}{FFC1B5}
\usepackage{threeparttable}
\usepackage{diagbox}
\usepackage{pifont}

\SetKwComment{Comment}{$\triangleright$\ }{}
\SetKwInput{KwInput}{Input}   
\SetKwInput{KwOutput}{Output} 
\SetKwInOut{Initialization}{Initialize} 

\ifCLASSINFOpdf
\else
\fi
\begin{document}

\title{Proto-Former: Unified Facial Landmark Detection by Prototype Transformer}
\author{Shengkai~Hu, Haozhe~Qi, Jun~Wan, Jiaxing~Huang, Lefei~Zhang,~\IEEEmembership{Senior Member, IEEE}, Hang~Sun, Dacheng~Tao,~\IEEEmembership{Fellow, IEEE}
	
    \thanks{This work is supported by the National Natural Science Foundation of
China (Grant No. 62571555  and 62002233), the Natural Science Foundation of Hubei Province, China (Grant No. 2024AFB992), the Fundamental Research Funds for the Central
Universities, Zhongnan University of Economics and Law (Grant No. 202511915). 
 Corresponding author: Jun Wan. }
    \thanks{ S. Hu, H. Qi, and J. Wan are with the School of Information Engineering, Zhongnan University of Economics and Law, Wuhan 430073, China, and with the School of Computer Science and Engineering (SCSE), Nanyang Technological University, Singapore 639798 (e-mail: shengkaihu@stu.zuel.edu.cn; hz777q@163.com; junwan2014@whu.edu.cn).}
    \thanks{L. Zhang is with the School of Computer Science, Wuhan University, Wuhan 430072, China (e-mail: zhanglefei@whu.edu.cn).}
    \thanks{H. Sun is with Hubei Key Laboratory of Intelligent Vision Based Monitoring for Hydroelectric Engineering, College of Computer and Information Technology, China Three Gorges University, Yichang 443002, China (e-mail: sunhang@ctgu.edu.cn).}
    \thanks{ J. Huang and D. Tao with the School of Computer Science and Engineering (SCSE), Nanyang Technological University, Singapore 639798. (e-mail: jiaxing.huang@ntu.edu.sg; dacheng.tao@ntu.edu.sg).}
	
}

\maketitle

\begin{abstract}Recent advances in deep learning have significantly improved facial landmark detection. However, existing facial landmark detection datasets often define different numbers of landmarks, and most mainstream methods can only be trained on a single dataset. This limits the model generalization to different datasets and hinders the development of a unified model. To address this issue, we propose Proto-Former, a unified, adaptive, end-to-end facial landmark detection framework that explicitly enhances dataset-specific facial structural representations (i.e., prototype). Proto-Former overcomes the limitations of single-dataset training by enabling joint training across multiple datasets within a unified architecture. Specifically, Proto-Former comprises two key components: an Adaptive Prototype-Aware Encoder (APAE) that performs adaptive feature extraction and learns prototype representations, and a Progressive Prototype-Aware Decoder (PPAD) that refines these prototypes to generate prompts that guide the model’s attention to key facial regions. Furthermore, we introduce a novel Prototype-Aware (PA) loss, which achieves optimal path finding by constraining the selection weights of prototype experts. This loss function effectively resolves the problem of prototype expert addressing instability during multi-dataset training, alleviates gradient conflicts, and enables the extraction of more accurate facial structure features. Extensive experiments on widely used benchmark datasets demonstrate that our Proto-Former achieves superior performance compared to existing state-of-the-art methods. The code is publicly available at: \href{https://github.com/Husk021118/Proto-Former}{https://github.com/Husk021118/Proto-Former}.
\end{abstract}


\IEEEpeerreviewmaketitle
\section{Introduction}
\IEEEPARstart{F}{acial} landmark detection (FLD), also known as face  alignment, has made great progress in recent years as a branch of computer vision. It aims to locate specific semantic facial landmarks, such as eyes, nose tip, mouth corners, etc. By accurately identifying facial landmarks, the geometric structure and pose information of the human face can be effectively captured, enabling a wide range of multimedia-oriented applications including dynamic facial expression recognition in video streams\cite{10878301_face_ana_tmm,liu2024confusable}, real-time avatar animation for virtual conferencing\cite{9974484_Real-time}, multimodal affective computing\cite{multimodal_affective_tmm,song2023srdf}, and enhanced face modeling for video-based content creation and editing\cite{10891424_tmm_liu,zhang2022information}.


\begin{figure}[t]
	\begin{center}
		\includegraphics[width=0.9\linewidth]{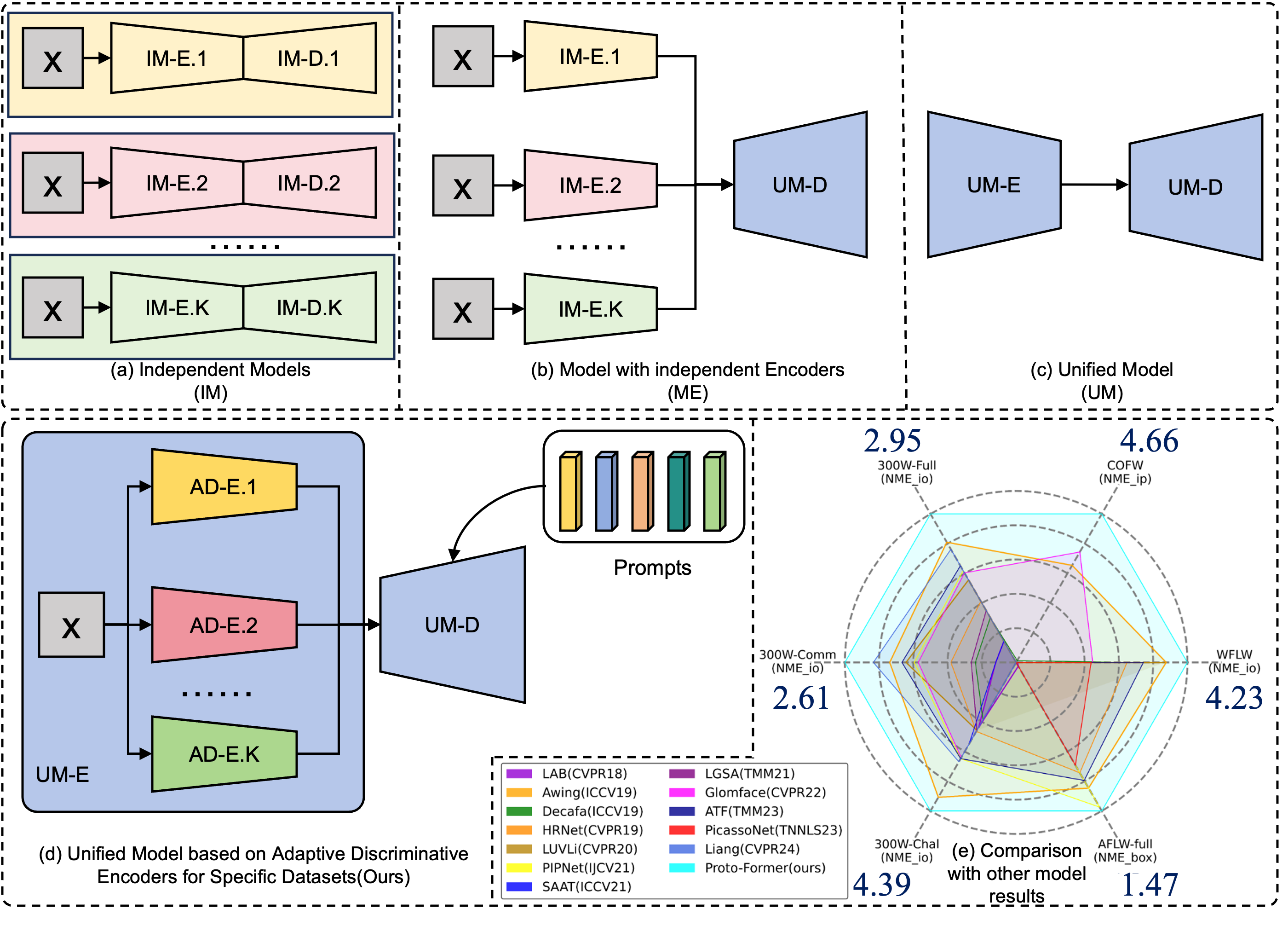}
	\end{center}
	\vspace{-1em}
	\caption{(a) Independent Model (IM), where IM.1, IM.2 and IM.K are independent. (b) Model with independent Encoders (ME), whose encoders (IM-E.1, IM-E.2 and IM-E.K) are independent. (c) Unified Model (UM), where UM-E and UM-D are the unified modules. (d) Our proposed unified Proto-Former is able to extract dataset-specific features similar to (a) by collaborating multiple adaptive discriminative encoders and an additional prompt block in the decoder. (e) Comparison with state-of-the-art FLD methods on four popular datasets 300W, COFW, WFLW, and AFLW.}
	\label{fig1}
	\vspace{-2em}
\end{figure}

With the advancement of deep learning, FLD methods based on CNN\cite{li2022towards,wan2020robust_1,wan2021robust_xi} and Transformers\cite{xia2022sparse,8653868_fld_tmm,wan2024precise_he} have made significant breakthroughs. However, they are still suffering from faces with large poses, severe occlusions or blur, because FLD datasets are relatively small in scale and cover limited complex scenarios, while the collection and annotation of new facial datasets is time-consuming and labor-intensive. Analyzing multiple FLD datasets, we found that while the number of landmarks varies across datasets (e.g., 68 in 300W, 19 in AFLW, and 98 in WFLW), they all describe facial structural information, and there are overlapping semantic landmarks among different datasets. Clearly, leveraging the overlapping semantic landmarks from other datasets can help improve the precision of landmark detection in more complex scenarios, while the unique landmarks can further enhance the modeling of facial structural information. These findings motivate our research into unified feature representation learning and unified facial structure modeling across multiple datasets.

In FLD, existing methods typically train independent models (IM) for each dataset (Fig.\ref{fig1}(a)) and have achieved favorable results \cite{wan2021robust,wing,wan2023precise_2}. These methods often require designing separate networks tailored to specific datasets for training and predicting a fixed number of landmarks, which hinders unified feature representation learning across multiple datasets and unified facial structure modeling.
Recently, all-in-one image restoration methods have been proposed to handle multiple degradation tasks. Some adopt multiple independent encoders with a shared decoder \cite{li2020all}(Fig.\ref{fig1}(b)): the encoders (IM-E.1, IM-E.2, IM-E.3) capture degradation-specific features, while the decoder (UM-D) aggregates them into a unified output. However, the disadvantage is that it is inefficient to use multiple independent encoders to process each degradation, and in practice the number of degradations is not fixed. Then, the unified model (Fig.~\ref{fig1}(c)) has been introduced\cite{zhang2025perceive,cui2024adair}, in which a shared encoder (UM-E) and decoder (UM-D) are employed to handle multiple degradation tasks. By jointly training on different degradation tasks, such models are able to capture a broad range of feature distributions, thereby improving their generalization capabilities. Moreover, previous pose estimation study
\cite{jeong2025posebh} has proposed multi-dataset joint training strategies using a shared encoder to effectively align heterogeneous datasets. Building upon this idea, the Mixture-of-Experts (MoE) mechanism \cite{fedus2022switchwxepert1,fan2022m3vit}, has further advanced unified models by dynamically activating experts for different feature representations, thereby alleviating gradient conflicts from cross-dataset distribution disparities and enhancing generalization within a unified framework. These developments provide new insights for us to design a novel unified FLD model for multiple datasets.


\begin{figure*}[t]
	\begin{center}
		\includegraphics[width=0.85\linewidth]{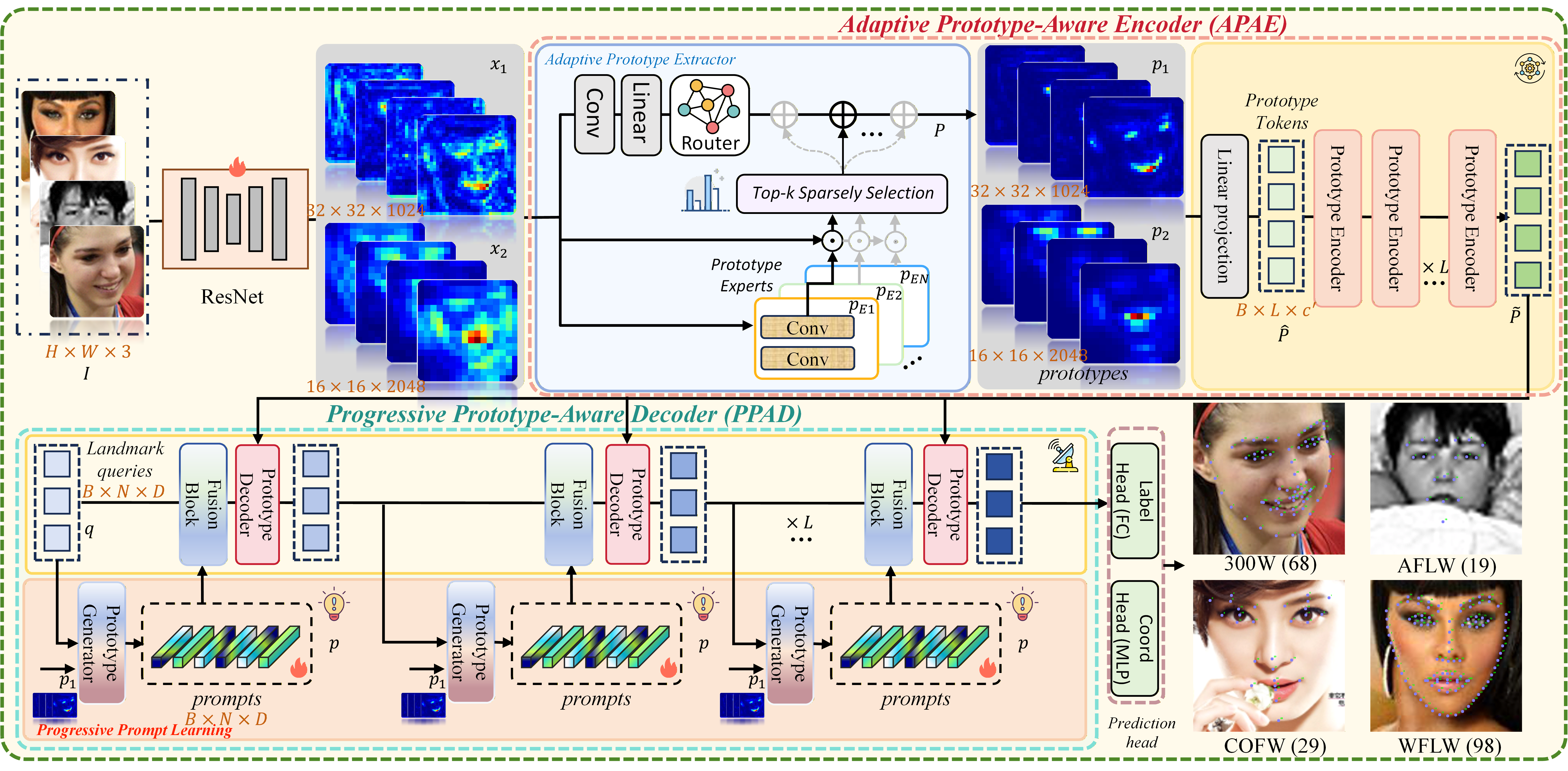}
	\end{center}
	\vspace{-1em}
	\caption{The network structure of the proposed Proto-Former. The image is processed through the backbone and APAE to handle features from different datasets. Reshaped features are fed into to the PPAD for dimensional information extraction, which enables High-resolution prototypes and landmark queries to be combined in the Prompt Generator to produce and inject prompts into the Proto-Decoder. Finally, the landmark query is used to predict the unified landmark index and its corresponding coordinates via the FC layer and the MLP layer.}
	\label{stru}
	\vspace{-1.5em}    
\end{figure*}

This paper proposes an adaptive, end-to-end, unified FLD model (i.e., Proto-Former as shown in Fig.\ref{fig1}(d)), which integrates multi-dataset training into a unified framework. Proto-Former can simultaneously predict varying numbers (e.g., 19, 29, 68, 98 or 124) of facial landmarks while significantly improving training efficiency and landmark detection  accuracy. The framework incorporates the Adaptive Prototype-Aware Encoder (APAE) and Progressive Prototype-Aware Decoder (PPAD). 
APAE aims to achieve adaptive perception of facial structure (i.e., prototypes) and then deeply models the prototype through the MHSA mechanism, thereby improving the model's ability to cope with the diversity of multiple datasets.
PPAD integrates a progressive landmark learning strategy, which uses the prototypes learned by APAE to guide interactions with dataset-specific features and global information, thereby better focusing on key facial regions and improving landmark detection accuracy.
In addition, a Prototype-Aware loss is proposed to guide optimal pathfinding in the dynamic routing space, enabling dataset-specific feature extraction and high-precision landmark detection.
Thus, our Proto-Former presents a significant advance in FLD (as shown in Fig.\ref{fig1}(e)). The main contributions of this work are summarized as follows:

1)  We propose the Proto-Former model integrates two innovative modules: APAE and PPAD. The APAE  is introduced to capture refined prototypes through Adaptive Prototype Extractor and MHSA mechanism, thereby addressing challenges such as inconsistent distributions across multiple datasets. Meanwhile, the PPAD leverages a progressive prompts learning strategy to deeply fuse prompt with the landmark queries, enhancing the model’s sensitivity to facial structure features.

2) A prototype-aware loss function is proposed to impose constraints on the activation distribution of the prototype expert to prevent the activations from being too dispersed, thereby alleviating the gradient conflicts caused by the unstable activations and multi-dataset training.


3) Our Proto-Former achieves state-of-the-art performance compared with state-of-the-art methods in four popular datasets: 300W, COFW, WFLW, AFLW. Notably, despite being based on coordinate regression, its accuracy surpasses that of most heatmap-based methods, demonstrating its robustness and effectiveness across multiple datasets.

The rest of this paper is organized as follows: Section \textbf{II} introduces the related work on FLD. Section \textbf{III} introduces the Proto-Former model, including the APAE  and APAD and the PA loss. Section \textbf{IV} evaluates the performance of Proto-Former through a large number of experiments. Finally, Section \textbf{V} gives the conclusions of this paper.

\section{Related Work}
FLD can be traced back to the end of the 19th century. Early FLD methods was template-based methods such as active shape models (ASM)\cite{cootes1995active}, constrained local models (CLM)\cite{cristinacce2006feature}, and random forest-based methods\cite{luo2015locating}. However, these methods have low model robustness and are sensitive to faces with pose variations. With the development of deep learning, a series of deep learning-based FLD methods have been proposed, which can be divided into two categories: heatmap regression and coordinate regression methods.

\textbf{Heatmap Regression methods.}
This kind of method regresses landmark heatmap and represents the position of each landmark as the peak of a two-dimensional Gaussian distribution. Heatmap regression methods can learn the spatial location distribution of landmarks in an image and are therefore more robust to pose, occlusion, and illumination variations. Dong et al.\cite{dong2018style} propose a style-aggregated approach to address the problems caused by the inherent differences in face images due to different image styles (such as grayscale and color images, bright and dark, strong contrast and soft contrast, etc.). Yang et al.\cite{yang2021transpose} propose a stacked hourglass network model to enhance the regression capability of the model. Huang et al.\cite{huang2021adnet} combine anisotropic direction loss (ADL) and anisotropic attention module (AAM) to improve its robustness. In order to solve the problem of semantic ambiguity, Wan et al.\cite{wan2021robust} propose a MMDN model, which improves the performance in complex scenes  by introducing multi-order feature associations and global shape constraints. Zhou et al.\cite{zhou2023star}propose the star loss, which uses the characteristics of semantic ambiguity to adjust and optimize, thereby reducing the impact of ambiguity on detection performance. Xiang et al.\cite{xiang2025popos} propose a POPoS framework, which leverages pseudo-range multilateration and a specially designed multilateration anchor loss to effectively correct heatmap errors and mitigate local optimum issues. Zhou et al.\cite{dang2025cascaded} propose a FLD method based on vision transformers, effectively modeling the geometric relationships among landmarks and enhancing feature propagation with its proposed Dual Vision Transformer (DViT) and Long Skip Connections (LSC). Heatmap regression methods rely on generating heatmaps to predict the locations of landmarks, which makes their performance limited by the resolution and scale of the generated heatmaps.

\textbf{Coordinate Regression Methods.}
Unlike heatmap regression methods, coordinate regression methods directly regress the coordinate values of landmarks. The advantage of such methods is that they are computationally efficient since they avoid the steps of generating and processing heatmaps. Although the performance of the coordinate regression method may not be as good as the heatmap regression method, it is usually suitable for application scenarios with high real-time requirements.
Feng et al.\cite{wing} introduce the Wing loss function, data augmentation strategy via a two-stage framework, improving robustness to faces with large head poses. Gao et al.\cite{tmm9082841} propose a coarse-to-fine FLD method with a landmark-guided self-attention (LGSA) module, enhancing global context and landmark focus, supported by an attentional consistency loss and a channel transformation block.
Xia et al.\cite{xia2022sparse} propose the SLPT model, which generates representations of each landmark from local image patches and aggregates these representations through attention mechanism, thereby learning more effective facial shape constraints and improving  landmark detection accuracy. 
Li et al.\cite{li2022towards} formulate FLD as a coordinate regression task, based on cascaded Transformer with a parallel decoder for more accurate FLD. 
Lan et al.\cite{9749863_atf_tmm} propose an Alternating Training Framework (ATF) that exploits inter-dataset commonalities and discrepancies under a weakly supervised paradigm, thereby enhancing the robustness and generalization of FLD across diverse annotation protocols. However, the performance of these methods is still limited by the scale of the dataset.

So far, on one hand, many researchers have focused on improving the model's localization ability within a single dataset, which hinders the generalization of face alignment models to different data distributions. On the other hand, although Transformer-based architectures are powerful, they often suffer from issues such as information redundancy and difficulty in focusing on task-relevant regions. To address these challenges, we draw inspiration from DETR\cite{carion2020end} and MoE\cite{yang2024multi} and propose an adaptive, end-to-end model for unified FLD. Additionally, we incorporate a novel prompt learning mechanism that enhances the model's ability to adaptively extract and utilize dataset-specific features by leveraging multi-dataset training and explicitly guiding the attention process through prompt learning. As a result, our approach surpasses the performance of state-of-the-art FLD methods.

\section{Method}
In this section, the definition of UFLD is given in \textbf{Section III. A}. Then, we present our proposed APAE in \textbf{Section III. B}, followed by the introduction of our proposed PPAD in \textbf{Section III. C}. Finally, \textbf{Section III. D} presents the proposed the PA loss.

\subsection{Unified Facial Landmark Detection (UFLD)}

UFLD refers to a new task aimed at jointly training a unified model using multiple datasets containing different number of landmarks, and being able to accurately predict the location of dataset-specific landmarks. However, implementing UFLD poses three major challenges. First, how to unify landmark definitions across different datasets, especially when the number and semantics of landmarks vary greatly. This requires a mechanism to combine dataset-specific landmarks into a common representation while maintaining accuracy and adaptability. The second is how to separate unified landmarks into dataset-specific landmarks during the training phase. Finally, how to resolve feature trends and gradient conflicts during multi-dataset training. Variations in data distribution, landmark definitions, and dataset scales can lead to gradient conflicts, which can negatively affect model convergence and performance.

\begin{figure}[h]
\begin{center}
	\includegraphics[width=0.9\linewidth]{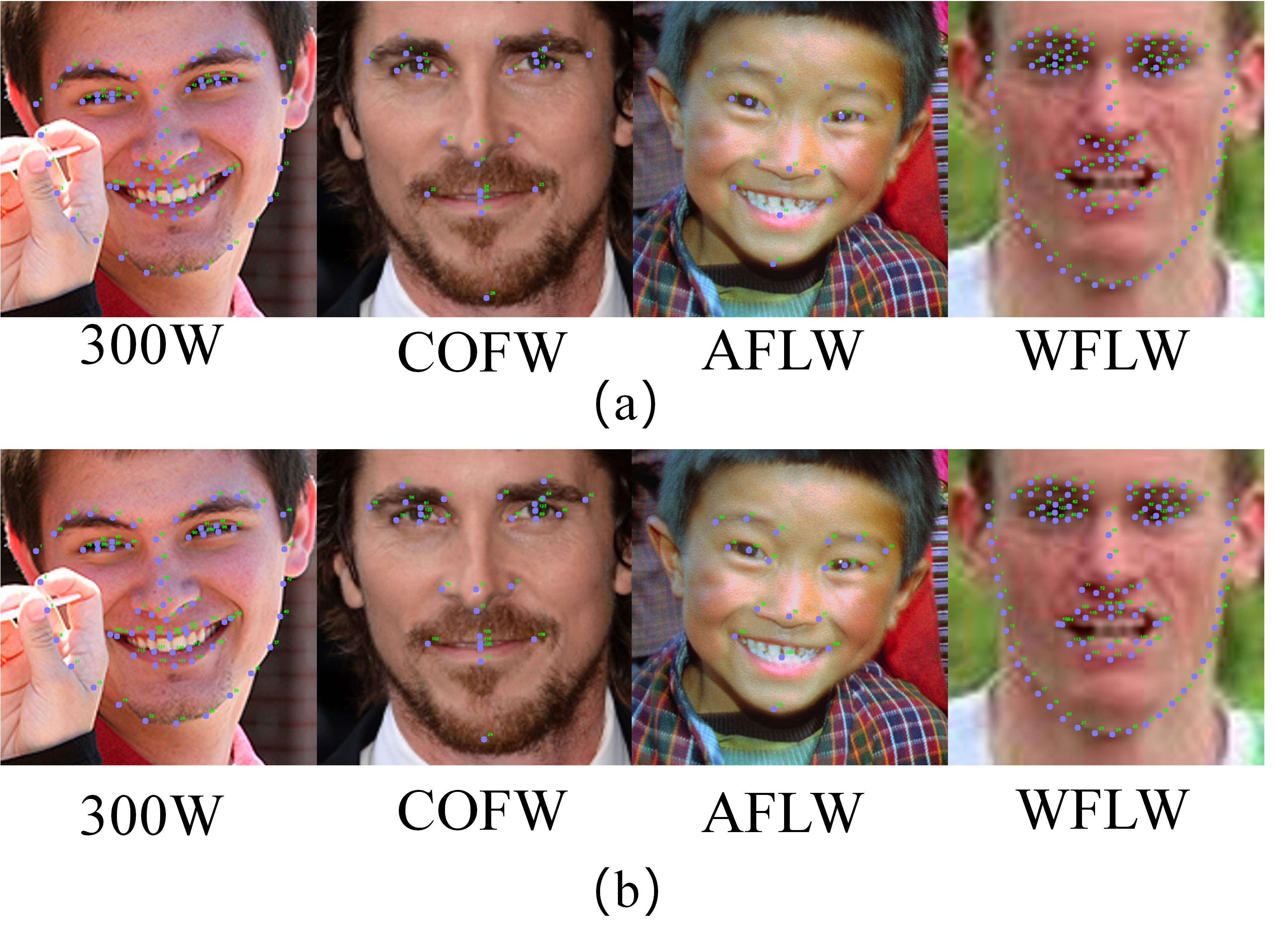}
	\end{center}
	\vspace{-2em}
	\caption{(a) The original landmark index. (b) the proposed unified landmark index. To achieve unified facial landmark detection, we combined the original landmarks from four popular datasets into 124 unified landmarks.}
	\label{mapping}
	\vspace{-1em}    
\end{figure}

\subsubsection{Unified Landmark Index} To address the problem of inconsistent landmark numbers and definitions across different datasets, we proposed a unified landmark version that integrates annotation information from popular datasets, such as 300W, WFLW, COFW, and AFLW. Fig.\ref{mapping} shows the unified and original landmark definitions. By assigning specific indexes to each facial landmark of multiple datasets,  semantic consistency and unified reference of landmarks can be achieved. And, we finally get 124 unified landmarks with clear semantics. The unified landmark index can realize the sharing of annotation information between multiple datasets, thereby improving the accuracy of landmark detection and forming a general framework that adapts to multiple datasets.

\subsubsection{Unified Landmark Matching}
To obtain dataset-specific landmark predictions from unified landmark predictions, we introduce the Hungarian algorithm \cite{carion2020end}. This algorithm selects relevant unified landmarks according to predefined indexes, separating dataset-specific landmarks from unified ones.

\subsubsection{Overall Architecture}
Given an input face image $\mathcal{I} \in R^{H \times W \times 3}$, where $H \times W$ denotes the spatial dimension. Proto-Former first extract multi-scale features $X= \{x_1,x_2| x_1\in R^{1024 \times 32 \times 32},x_2\in R^{2048 \times 16 \times 16}\}$ with the ResNet backbone. Then, $X$ undergoes the APAE, which contains an Adaptive Prototype Extractor (APE) and several Proto-Encoders. Specifically, the APE will process the features $X$ into multi-scale prototype $P$ with the same dimension as $X$. $P$ will be transformed into sequence representations respectively and stacked together to obtain $\hat{P}$. $\hat{P}$ will be fed into a 6-level Proto-Encoders and output refined prototype $\tilde{P}$. In PPAD, a randomly initialized landmark query $q$ will be processed by the prompt generator to output prompt $p$, which will be combined with $q$ to obtain refined query $\hat{q}$ by a fusion block. $\hat{q}$ will be queried with refined prototype $\tilde{P}$ by a Proto-Decoder. The final output of Proto-Decoder $q_L$ will be processed by a MLP layer and a Linear layer to obtain the landmark coordinate predictions and landmark label index. Furthermore, we also propose PA loss to guide the learning of the prototypes. Next, we describe the proposed APAE, PPAD and PA loss in detail.

\subsection{Adaptive prototype-Aware Encoder (APAE)}
Different FLD datasets contain different numbers of landmarks, posing significant challenges to developing robust and unified models. In addition, different FLD datasets also exhibit different characteristics. For example, the 300W dataset focuses on the frontal face and covers a wide range of age groups, while the COFW dataset focuses on heavily occluded faces. The AFLW dataset contains different viewpoints, while the WFLW dataset emphasizes rich facial expressions and variations. These differences also needed to be distinguished and learned by the unified model. 

To address these challenges, we propose an APAE which consists of a APE and serval Prototype Encoders (Proto-Encoder). APE aims to construct a dynamic routing space consisting of multiple prototype experts, each of which is responsible for processing part of the facial structure. The Proto-Encoder uses the MHSA mechanism to deeply model and enhance the prototype, assisting the decoder in reasoning the dataset-specific landmarks by capturing the hidden contextual associations and feature hierarchical relationships in the dataset.

\subsubsection{Adaptive Prototype Extractor}
APE dynamically selects the TopK prototype experts through the routing mechanism and then combines their outputs with corresponding gating scores to produce the prototype $P$. 
The  whole process can be defined as:
\begin{equation}
    \ p_1 = [\sum_{k=1}^{K} (g_k\cdot {\mathcal{P}_k(x_1)})]\odot x_1.
\end{equation} where $K$ denotes the number of prototype experts selected by the TopK function, $\mathcal{P}_{k}$ denotes the $k$-th selected prototype expert, $g_k \in \mathcal{G}$ denotes the corresponding gating score. We can also obtain another scale of prototype $p_2$ used the similar operation and $P= \{p_1,p_2| p_1\in R^{1024 \times 32 \times 32},p_2\in R^{2048 \times 16 \times 16}\}$.

\textbf{Prototype Expert.} Due to significant differences in facial attributes and imaging conditions, UFLD across different datasets faces unique challenges. To address this issue, we introduce prototype experts that can encode the characteristics of different datasets and focus on regional facial features. The prototype expert can be achieved by using low-rank decomposition. 

Specifically, two convolutional layers are used for low-rank transformation. The first $3 \times 3$ convolution reduces the channel dimension from $d_{i}$ to a smaller rank $d_{r}$, thereby capturing essential features and discarding redundancy. Another $1 \times 1$ convolution increases the dimension back to $d_{o}$, reconstructing the output features with minimal information loss, where $d_{r} \ll d_{o}$:

\begin{equation}
	\mathcal{P}_{E}=(Conv_B\cdot Conv_A) \cdot x+b
\end{equation}where $B$ and $A$ corresponds to the above two convolution operations, $b$ denotes the bias. 


\begin{figure}[t]
\begin{center}
	\includegraphics[width=0.85\linewidth]{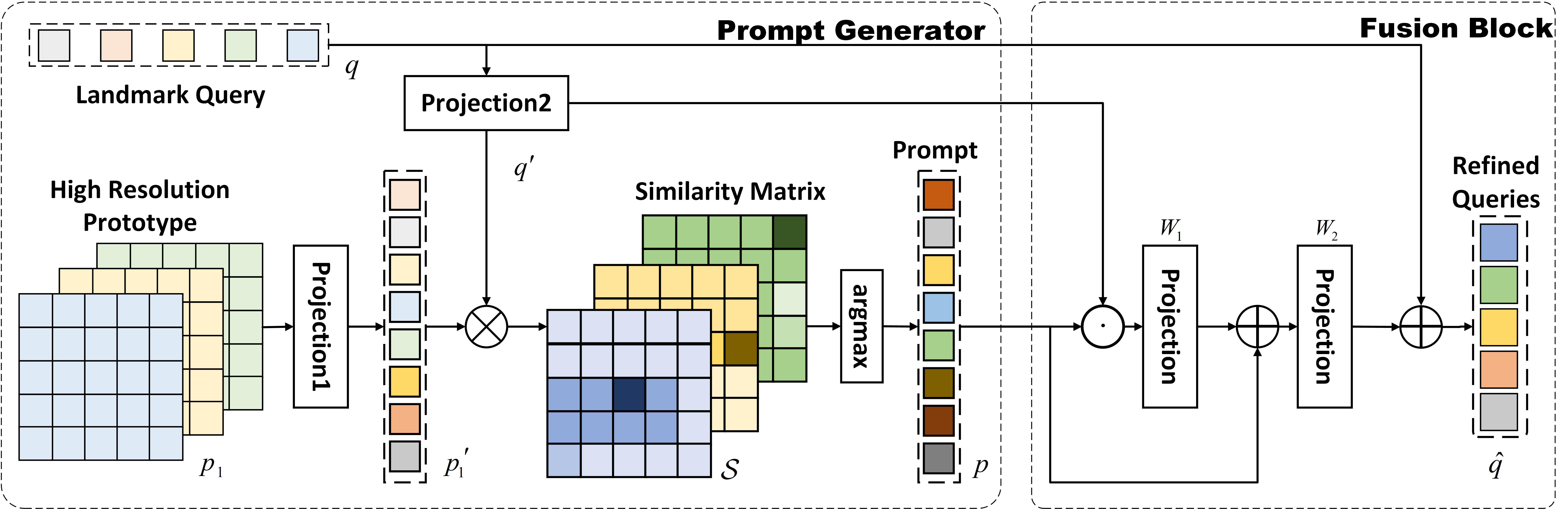}
	\end{center}
	\vspace{-1em}
	\caption{The proposed prompt generator and fusion block primarily consist of a prompt generation mechanism based on the similarity matrix and a fusion process that integrates the original landmark queries with the generated prompts. This design effectively enhances the model’s attention to facial structural features, leading to improved performance.}
	\label{prompt_encoder}
	\vspace{-2em}    
\end{figure}

 \textbf{Routing Mechanism.} The routing mechanism is the key to dynamic addressing, directly determining whether the model can select effective prototypes. To enhance the model’s ability to perceive facial structures under complex scenarios, we employ Multi-Head Self-Attention (MHSA) and Position Awareness block to assist feature extraction and implement dynamic routing. This routing mechanism consists of three steps: feature extraction, feature distribution estimation, and expert index generation.

By inputting $X$, the routing mechanism used a $3\times3$ convolution layer and a reshape operation to obtain the feature sequence $S \in R^{\frac{C}{4} \times WH}$. Then the MHSA is applied to capture the hidden contextual associations and feature hierarchical relationships in the dataset, which can be formulated as follows: 
\begin{equation}
	S'=MHSA( W_Q\cdot S, W_K \cdot S, W_V\cdot S)
\end{equation}
where $W_Q$, $W_K$, $W_V$ are corresponding the mapping matrix. Inspired by \cite{hou2020strip}, the position awareness block is also introduced to enhance the long-range spatial context location information. The position awareness block contains two MLPs, one of which transforms $R^{HW \times C}$  into $R^{HW \times 1}$  along the channel dimension, and the second MLP further transforms $R^{HW \times 1} $ into $R^{HW \times N} $ along the channel dimension. After that, it will be concatenated $S'$
along channel dimension, followed by a $1\times1$ convolution layer to reduce the channel dimension to $N$. Finally, the softmax activation $\sigma$ is applied to calculate the gating scores $G$, and the TopK function will be further applied to generate indexes of activated experts:
\begin{equation}
	{\mathcal{I}}= TopK(G,K), \mathcal{G}= G[{\mathcal{I}}]
\end{equation}where $G \in R^{1 \times N} $ denotes the probability of N experts, $\mathcal{I}\in R^{1 \times K}$ denotes the TopK expert indexes, which will then apply a broadcast in $G$ to generate the selected TopK prototype expert gating scores $\mathcal{G} \in R^{1\times K}$. The selected TopK prototype experts are combined through their corresponding gating scores $\mathcal{G}$ to generate the prototype $P$, which is then fed into the Proto-Encoders.

\subsubsection{Prototype Encoder}
Proto-Encoder is used to capture the hidden contextual associations and hierarchical feature  relationships in the dataset. Given the prototypes $p_1 \in R^{1024 \times 32 \times 32}$ and $p_2 \in R^{2048 \times 16 \times 16}$ generated by the APE, they are first processed to align their channel dimensions to $c’$. Subsequently, both are transformed into sequences and concatenated along the spatial dimension, resulting in tokens $\hat{P} \in R^{l \times c’}$, where $l = 32 \times 32 + 16 \times 16$. These tokens, $\hat{P}$, are then passed through $L$ Proto-Encoder blocks. The outputs from these blocks are fused using an MLP layer and scaled by a hyperparameter to produce the refined prototypes $\tilde{P}$. This process can be defined as:
\begin{equation}
	\tilde{P}=\lambda \cdot MLP(cat[ \mathbb{E}_{1}(\hat{P}_0), ...,\mathbb{E}_{L-1}(\hat{P}_{L-2})])+ \mathbb{E}_L(\hat{P}_{L-1})
\end{equation}
where $\mathbb{E}(\cdot)$ denotes the Proto-Encoder. These refined prototypes are treated as the Value of MHCA in next PPAD.







   




\subsection{Progressive Prototype-Aware Decoder (PPAD)}
In many current studies\cite{carion2020end,zhu2020deformable}, queries were typically not specifically enhanced after initialization to emphasize key region features. Inspired by \cite{watchareeruetai2022lotr}, we propose an innovative PPAD, which includes multiple prompt generators, fusion blocks and Proto-Decoders, as shown in Fig.\ref{prompt_encoder}. The prompt generator aims to generate prompts and fuse them with landmark queries to enhance Proto-Decoder's query capability for key region features. By cascading multiple prompt generators in a progressive prompt learning manner, the PPAD iteratively refines the prompts, enabling more effective landmark queries. This, in turn, facilitates the detection of landmarks with higher accuracy.

\subsubsection{Prompt Generator}
The prompt generator refines landmark queries by leveraging the similarity matrix between the prototypes $p_1$  and the landmark queries  $q$. This process selects the most relevant features of facial structural components as prompts, which are then fused with the landmark queries from the previous layer. The resulting refined landmark queries serve as guidance for subsequent processing, enabling more accurate landmark detection.

Given the high-resolution prototype $p_1\in R^{1024\times\ 32\times\ 32}$ generated by APE, it will be first processed by the $projection1$ operation. At the same time, landmark queries undergo another $projection2$ operation. These operations align the landmark queries and prototype in the channel dimension, resulting in  $p_1'\in R^{HW\times D}$  for the prototype and  $q'\in R^{N\times D}$  for the landmark queries, where  $N$  and  $D$  represent the number of predefined landmark queries and channel dimension, respectively. Then we calculate the similarity matrix $S$ between $p_1'$ and $q'$,
and the argmax operation is performed on $S$ along the channel dimension to obtain the indexes of prompts $G=argmax(S)$, and $G \in R^{1 \times N}$. After that the prompt corresponding to the indexes will be selected from the high-resolution prototype $p'_1$, which can be defined as:
\begin{equation}
	p=p_1'[G]
\end{equation}
where $p \in R^{N \times D}$ denotes the selected prompts, which will then fed into fusion block for obtaining  refined landmark queries $\hat{q}$.
\subsubsection{Fusion Block}
Inspired by \cite{mehta2021mobilevit} , the fusion block first uses the computationally efficient element-wise product to implement the interaction between $q'$ and $p$, and then adjusts the result through a projection layer (i.e., $W_1$). The process is defined as:
\begin{equation}
	\mathcal{A}=(q'\odot p)\cdot W_1
\end{equation}
where $W_1\in R^{D \times D}$ denotes the matrix corresponding to the above projection layer and $\mathcal{A}$ denotes the obtained result. After that, a learnable parameter $\alpha \in R^{1 \times D}$ is used to re-weight the normalized $\mathcal{A}$, which will then be added back to the selected prompt $p$. To further refine the landmark queries, a projection operation (i.e., $W_2$) followed by a residual connection is also employed. The whole process can be defined as:
\begin{equation}
	\hat{q}=q+(\alpha \odot \frac{\mathcal{A}}{\|\mathcal{A}\|}_2+p)\cdot W_2 
\end{equation}
where $\hat{q}$ denotes the refined landmark queries and $\hat{q} \in R^{N\times D}$. ${\| \cdot \|}_2$ denotes $L_2$ normalization operation. $\hat{q}$ will then be fed into the Proto-Decoder to assist the decoding process. 

\subsubsection{Prototype Decoder}
The Proto-Decoder aims to predict the coordinates of facial landmarks and their corresponding label indexes by interacting between the refined landmark queries $\hat{q}$ and the prototypes $\tilde{p}$.

Assuming $\hat{q}_{i-1}$ denotes the output of the previous Proto-Decoder, the multi-head attention mechanism can be calculated as:
\begin{equation}
    \tilde{q}_i=\hat{q}_{i-1} + LN(MHCA(\hat{q},q,q))
\end{equation}
\vspace{-1em}
\begin{equation}
    \tilde{q}_i'=MHCA(\tilde{q}_i,\tilde{P},\tilde{P})
\end{equation}
\vspace{-1em}
\begin{equation}
\hat{q}_i=FFN(LN(\tilde{q}_i') + \tilde{q}_i)
\end{equation}
where $LN$ denotes the LayerNorm operation, $\tilde{P}$ denotes the refined prototypes, $FFN$ denotes the Feed-Forward Network.

The output of PPAD $q_L$ will be processed by the prediction head to obtain the unified landmark label indexes prediction $O^{index}\in R^{N\times (124+1)} $ and landmark coordinate prediction $O^{coord}\in R^{N\times 2}$. $+1$ means the model predicts an additional ``no landmark'' category in case the embedding does not correspond to any landmark. To obtain dataset-specific landmark predictions, we can use the predefined unified landmark index.

\subsection{Prototype-Aware Loss}
To leverage the characteristics of different datasets, a multi-dataset joint training strategy is used for improving the landmark detection accuracy. However, this strategy introduces new challenges, such as gradient conflicts across datasets and instability in expert assignment.
From the t-SNE analysis results corresponding to the feature maps processed by the backbone (as shown in Fig.\ref{tsne} (a)), it can be seen that there is no significant difference in the distribution of samples from different datasets, and it is impossible to clearly distinguish them in t-SNE. To address these issues, we incorporate a novel supervisory signal, namely, Prototype-Aware (PA) loss, designed to stabilize the expert routing and selecting within the APAE. The PA loss learns prototypes by aligning the expert distributions of samples within the same dataset. Specifically, for two samples within a batch with different gating scores, their similarity is computed as follows:
\begin{equation}
	s_{ij} = \frac{s_i \cdot s_j}{\|s_i\| \cdot \|s_j\|}
\end{equation}
where $s_i$ and $s_j$ denote the $i$-th sample's gating scores and the $j$-th sample's gating scores within one mini-batch, respectively. $\|\cdot\|$ denotes the Euclidean norm. Therefore, the PA loss can be calculated as:
\begin{equation}
\mathbb{L}_{PA}= \sum_{i=1}^{B-1} \sum_{j=i+1}^{B} \left( 1 - s_{ij} \right)
\end{equation}
where $B$ denotes the number of samples in a batch. The proposed PA loss reduces the expert selection differences of samples within the same dataset, while increasing the expert selection differences across datasets. This approach effectively alleviates the gradient conflict and expert assignment instability between datasets, promotes the learning of prototype features and realizes a unified framework for FLD.

To address the FLD task, we introduce a landmark coordinate loss $\mathbb{L}_{coor}$ (an ${\ell_1}$ loss) and a landmark index loss $\mathbb{L}_{index}$ (a cross-entropy loss). The overall loss can be defined as:
\begin{align}
	\mathbb{L} = \lambda_1 \mathbb{L}_{coor}  + \lambda_2 \mathbb{L}_{index} + \lambda_3 \mathbb{L}_{PA} 
\end{align}
where $\lambda_1$, $\lambda_2$, and $\lambda_3$ balance the contributions of each loss term.

\section{Experiments}
In this section, we introduce the evaluation metrics on popular datasets. We conduct experiments on four popular datasets (300W\cite{300w}, COFW\cite{cofw}, WFLW\cite{wflw}, and AFLW\cite{aflw}) and show the comparison results between our method and the SOTA FLD method. Finally, we perform ablation studies on the network components and evaluate their effectiveness.
\subsection{Dataset and Implementation details}
\textbf{300W} (68 landmarks)\cite{300w}: It is a commonly used face alignment dataset. There are 3148 images for training and 689 images for testing, which are annotated with 68 landmarks. The testset is further divided into common Subset and challenging Subset. The common Subset includes 224 images from the LFPW\cite{lfpw} testset and 330 images from the Helen testset. The challenging Subset\cite{ibug} comprises 135 images characterized by significant variations, posing greater difficulty for FLD algorithms. 

\textbf{WFLW} (98 landmarks)\cite{wu2018look}: The WFLW dataset contains 7,500 training images and 2,500 test images, each annotated with 98 facial landmarks. The test set is divided into several Subsets for specific variations. This detailed annotation and Subset division makes WFLW a comprehensive benchmark for robust FLD.

\textbf{COFW} (29 landmarks)\cite{cofw}: The COFW dataset is specifically designed to evaluate FLD models under heavy occlusion. It contains 1,345 face images, each annotated with 29 facial landmarks, including faces with various levels of occlusion caused by objects, hands, or accessories. Among these, 845 images are used for training, and  the remaining 500 images form the testset.

\textbf{AFLW} (19 landmarks)\cite{aflw}: The AFLW dataset contains 24,368 faces with significant pose variations, making it a reliable benchmark for FLD. Each face is annotated with up to 21 landmarks. To ensure a fair comparison with other methods \cite{awing,wing}, we follow the protocol in \cite{aflw} to reduce the annotations to 19 landmarks to ensure consistency in the evaluation.

\begin{figure}[t]
\begin{center}
	\includegraphics[width=0.9\linewidth]{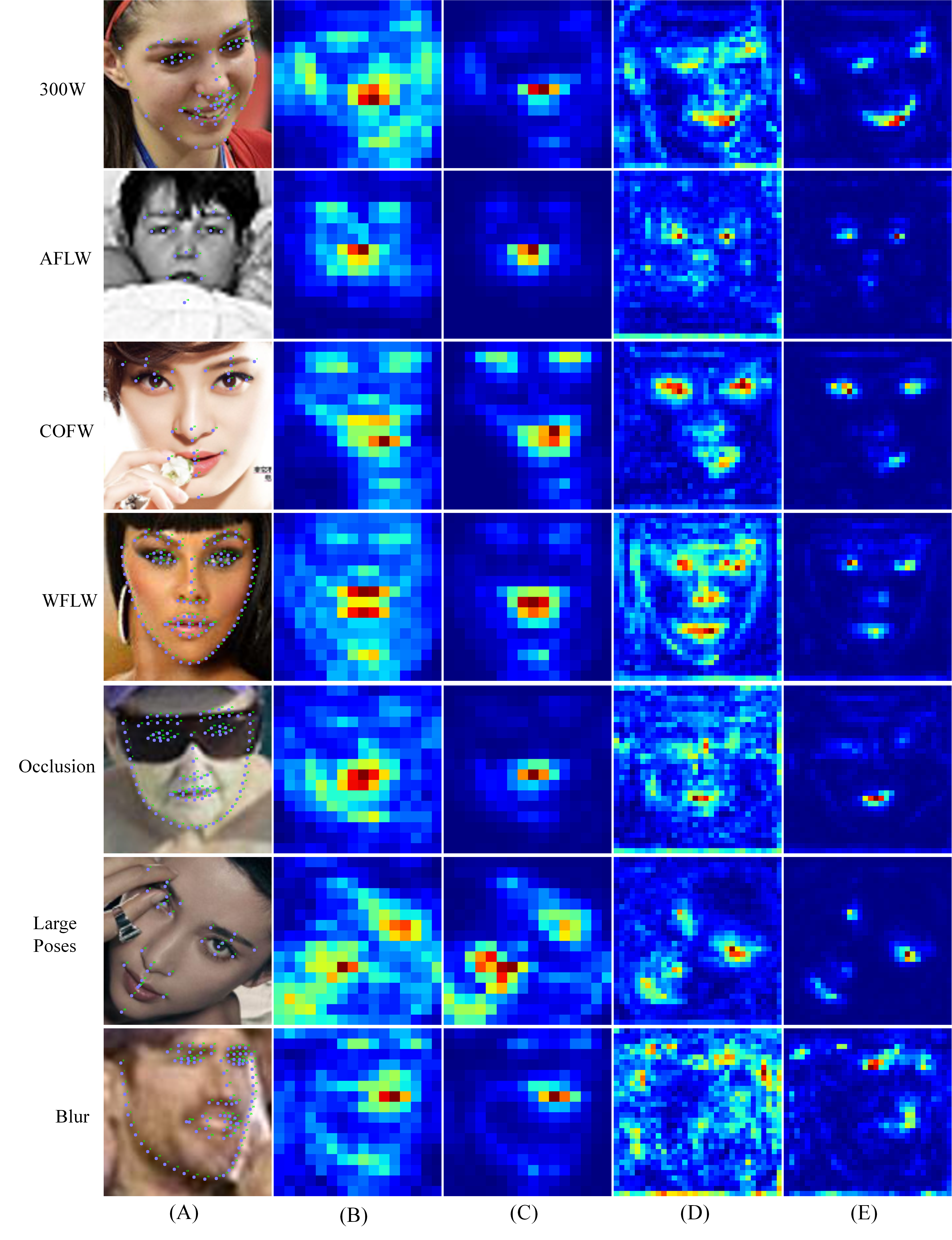}
	\end{center}
	\vspace{-1em}
	\caption{Comparison of prototypes and feature maps in normal (the first 4 rows) and complex circumstances (the last 3 rows). (a) predicted landmarks, (b) low-resolution feature map, (c) low-resolution prototype,  (d) high-resolution feature map and (e) high-resolution prototype. It demonstrates our proposed Proto-Former can extracts effective prototypes.}
	\label{exp_res1}
	\vspace{-1em}    
\end{figure}

\begin{table}[t]
    \centering
    \scriptsize
    \caption{Comparisons with SOTA methods on the 300W dataset. 
    The error (NME) is normalized by the inter-ocular distance. 
    $\circ$ and $\diamond$ denote heatmap regression and coordinate regression methods, respectively. (\% omitted)}
    \renewcommand\arraystretch{0.9}
    \vspace{-1.4em}
    \begin{tabular}{p{3.8cm}|ccc}
     \toprule[1pt]
        Method  & Common  & Challenging  & Full  \\
        \midrule
        \makebox[3.8cm][l]{$\circ$\, LAB (CVPR18) \cite{wu2018look}} & 2.98 & 5.19 & 3.49 \\
        \makebox[3.8cm][l]{$\circ$\, AWing (ICCV19) \cite{awing}} & 2.72 & 4.52 & 3.07 \\
        \makebox[3.8cm][l]{$\circ$\, LUVLi (CVPR20) \cite{kumar2020luvlifacealignmentestimating}} & 2.76 & 5.16 & 3.23 \\
        \makebox[3.8cm][l]{$\circ$\, SAAT (ICCV21) \cite{saatzhu2021improving}} & 2.82 & 5.03 & 3.25 \\
        \makebox[3.8cm][l]{$\circ$\, ADNet (ICCV21) \cite{huang2021adnet}} & 2.53 & 4.58 & 2.93 \\
        \makebox[3.8cm][l]{$\circ$\, STAR (CVPR23) \cite{zhou2023star}} & 2.52 & 4.32 & 2.87 \\
        \midrule
        \makebox[3.8cm][l]{$\diamond$\, ODN (CVPR19) \cite{odn}} & 3.56 & 6.67 & 4.17 \\
         \makebox[3.8cm][l]{$\diamond$\, DAG (ECCV20) \cite{dag}} & 2.62 & 4.77 & 3.04 \\
        \makebox[3.8cm][l]{$\diamond$\, LGSA (TMM21) \cite{tmm9082841}} & 2.92 & 5.16 & 3.36 \\
        \makebox[3.8cm][l]{$\diamond$\, PIPNet (IJCV21) \cite{pipnetJLS21}} & 2.78 & 4.89 & 3.19 \\
        \makebox[3.8cm][l]{$\diamond$\, SLPT (CVPR22) \cite{xia2022sparse}} & 2.75 & 4.90 & 3.17 \\
        \makebox[3.8cm][l]{$\diamond$\, GlomFace (CVPR22) \cite{glomfacezhu2022occlusion}} & 2.79 & 4.87 & 3.20 \\
        \makebox[3.8cm][l]{$\diamond$\, DTLD (CVPR22) \cite{DTLD}} & 2.59 & 4.50 & 2.96 \\
        \makebox[3.8cm][l]{$\diamond$\, ATF (TMM23) \cite{9749863_atf_tmm}} & 2.75 & 4.89 & 3.17 \\
        \makebox[3.8cm][l]{$\diamond$\, EfficentFan (TNNLS23) \cite{pipnetJLS21}} & 2.98 & 5.21 & 3.42 \\
        \makebox[3.8cm][l]{$\diamond$\, PicassoNet (TNNLS23) \cite{picassonet}} & 3.03 & 5.81 & 3.58 \\
        \makebox[3.8cm][l]{$\diamond$\, Lite-HRNet (ICIP23) \cite{Lite-HRNet}} & 3.97 & 6.89 & 4.54 \\
        \makebox[3.8cm][l]{$\diamond$\, Liang et al. (CVPR24) \cite{liang2024}} & 2.68 & 4.86 & 3.10 \\
        \hline
        \rowcolor{my_color}\makebox[3.8cm][l]{$\diamond$\, \textbf{Proto-Former (ours)}} & 
        \textbf{2.61} & \textbf{4.39} & \textbf{2.95} \\
     \bottomrule[1pt]
    \end{tabular}
    \vspace{-2em}
    \label{300wtable}
\end{table}

        

\begin{table}[t]
  \centering
  \scriptsize
  \caption{Comparisons with SOTA methods on the COFW dataset. 
  The error (NME) is normalized by the inter-pupil distance. 
  $\circ$ and $\diamond$ denote heatmap regression and coordinate regression methods, respectively. (\% omitted)}
  \vspace{-1.4em}
  \renewcommand\arraystretch{0.4}
  \begin{tabular}{p{4.0cm}|cc}
      \toprule[1pt]
        Method & $\rm NME_{ip}$  & FR (Failure Rate) \\
      \midrule
        \makebox[4.0cm][l]{$\circ$\, AWing (ICCV19) \cite{awing}} & 4.94 & 0.99 \\
        \makebox[4.0cm][l]{$\circ$\, SCPAN (TCYB21) \cite{li2023cascaded}} & 5.81 & 3.55 \\
        \makebox[4.0cm][l]{$\circ$\, STAR (CVPR23) \cite{zhou2023star}} & 4.62 & 0.79 \\
        
        \makebox[4.0cm][l]{$\circ$\, CIT-v2 (IJCV24) \cite{wan2024precise_he}} & 4.93 & 1.58 \\
        
      \midrule
        \makebox[4.0cm][l]{$\diamond$\, ODN (CVPR19) \cite{odn}} & 5.30 & - \\
        \makebox[4.0cm][l]{$\diamond$\, MMDN (TNNLS22) \cite{mmdn}} & 5.01 & 1.78 \\
        \makebox[4.0cm][l]{$\diamond$\, GlomFace (CVPR22) \cite{glomfacezhu2022occlusion}} & 4.37 & 1.56 \\
        \makebox[4.0cm][l]{$\diamond$\, SLPT (CVPR22) \cite{xia2022sparse}} & 4.79 & 1.18 \\
        \makebox[4.0cm][l]{$\diamond$\, DSLPT-R50 (TPAMI23) \cite{10079153}} & 4.81 & 1.18 \\
      \hline
        \rowcolor{my_color}\makebox[4.0cm][l]{$\diamond$\, \textbf{Proto-Former (ours)}} & 
        \textbf{4.67} & \textbf{0.20} \\
      \bottomrule[1pt]
  \end{tabular}
  \label{cofwtable}
  \vspace{-2em}
\end{table}


\begin{table*}[t]
  \centering
  \scriptsize
  \caption{Comparisons with SOTA methods on WFLW Subset. 
  NME is normalized by the inter-ocular distance. 
  $\circ$ and $\diamond$ denote heatmap regression and coordinate regression methods, respectively. (\% omitted).}
  \vspace{-1.5em}
  \renewcommand\arraystretch{0.5}
  \resizebox{\textwidth}{!}{%
  \begin{tabular}{p{2.8cm}|ccccccc}
      \toprule[1pt]
       Method
     & Testset & Pose Subset & Expression Subset & Illumination Subset & Make-Up Subset & Occlusion Subset & Blur Subset \\
      \midrule
        \makebox[2.8cm][l]{$\circ$\, LAB(CVPR18) \cite{wu2018look}} & 5.27 & 10.24 & 5.51 & 5.23 & 5.15 & 6.79 & 6.32 \\
        \makebox[2.8cm][l]{$\circ$\, Wing(CVPR18) \cite{wing}} & 4.99 & 8.75 & 5.36 & 4.93 & 5.41 & 6.37 & 5.81 \\
        \makebox[2.8cm][l]{$\circ$\, DeCaFA(ICCV19) \cite{DeCaFA}} & 4.62 & 8.11 & 4.65 & 4.41 & 4.63 & 5.74 & 5.38 \\
        \makebox[2.8cm][l]{$\circ$\, HRNet(CVPR19)\cite{hrnetSunXLW19}} & 4.60 & - & - & - & - & - & - \\
        \makebox[2.8cm][l]{$\circ$\, AWing(ICCV19)\cite{awing}} & 4.36 & - & - & - & - & - & - \\
        \makebox[2.8cm][l]{$\circ$\, SCPAN (TCYB21) \cite{li2023cascaded}} & 4.29 & 7.22 & 4.68 & 4.34 & 4.21 & 5.25 & 4.88 \\
         
      \midrule
         \makebox[2.8cm][l]{$\diamond$\, DAG(ECCV20)\cite{dag}} & 4.21 &7.36 & 4.49 & 4.12 & 4.05 & 4.98 & 4.82 \\
        \makebox[2.8cm][l]{$\diamond$\, PIPNet(IJCV21)\cite{pipnetJLS21}} & 4.31 & - & - & - & - & - & - \\
        \makebox[2.8cm][l]{$\diamond$\, MMDN(TNNLS22)\cite{mmdn}} & 4.87 & 7.71 & 4.79 & 4.61 & 4.72 & 6.17 & 5.72 \\
        \makebox[2.8cm][l]{$\diamond$\, GlomFace(CVPR22)\cite{glomfacezhu2022occlusion}} & 4.81 & 8.71 & - & - & - & 5.14 & - \\
        \makebox[2.8cm][l]{$\diamond$\, DTLD(CVPR22)\cite{DTLD}} & 4.08 & - & - & - & - & - & - \\
        \makebox[2.8cm][l]{$\diamond$\, ATF(TMM23)\cite{9749863_atf_tmm}} & 4.50 & 7.54 & 4.65 & 4.45 & 4.20 & 5.30 & 5.19 \\
        \makebox[2.8cm][l]{$\diamond$\, EfficentFan(TNNLS23)\cite{efficientfan}} & 4.54 & 8.20 & 4.87 & 4.39 & 4.54 & 5.42 & 5.04 \\
        \makebox[2.8cm][l]{$\diamond$\, PicassoNet(TNNLS23)\cite{picassonet}} & 4.82 & 8.61 & 5.14 & 4.73 & 4.68 & 5.91 & 5.56 \\
        \makebox[2.8cm][l]{$\diamond$\, Lite-HRNet(ICIP23)\cite{Lite-HRNet}} & 5.58 & 9.79 & 6.13 & 5.44 & 5.87 & 6.57 & 6.05 \\
      \hline
        \rowcolor{my_color}\makebox[2.8cm][l]{$\diamond$\, \textbf{Proto-Former (ours)}} & 
        \textbf{4.23} & \textbf{7.09} & \textbf{4.44} & \textbf{4.22} & \textbf{4.08} & \textbf{5.00} & \textbf{4.94} \\
      \bottomrule[1pt]
  \end{tabular}%
  }
  \label{tabwflw}
  \vspace{-2em}
\end{table*}

\begin{table}[t]
  \centering
  \scriptsize
  \caption{Comparisons with SOTA methods on the AFLW dataset. 
  The error (NME) is normalized by face size. 
  $\circ$ and $\diamond$ denote heatmap regression and coordinate regression methods, respectively. (\% omitted)}
   \vspace{-1.5em}
  \renewcommand\arraystretch{0.5}
  \begin{tabular}{p{5.0cm}|c}
      \toprule[1pt]
        Method & Testset \\
      \midrule
        \makebox[5.0cm][l]{$\circ$\, LAB (CVPR18) \cite{wu2018look}} & 1.85 \\
        \makebox[5.0cm][l]{$\circ$\, HRNet (CVPR19) \cite{hrnetSunXLW19}} & 1.57 \\
        \makebox[5.0cm][l]{$\circ$\, AWing (ICCV19) \cite{awing}} & 1.53 \\
        \makebox[5.0cm][l]{$\circ$\, LUVLi (CVPR20) \cite{kumar2020luvlifacealignmentestimating}} & 2.28 \\
        \makebox[5.0cm][l]{$\circ$\, SCPAN (TCYB21) \cite{li2023cascaded}} & 2.01 \\
      \midrule
        \makebox[5.0cm][l]{$\diamond$\, PIPNet (IJCV21) \cite{pipnetJLS21}} & 1.48 \\
        \makebox[5.0cm][l]{$\diamond$\, DTLD (CVPR22) \cite{DTLD}} & 1.38 \\
        \makebox[5.0cm][l]{$\diamond$\, ATF (TMM23) \cite{9749863_atf_tmm}} & 1.55 \\
        \makebox[5.0cm][l]{$\diamond$\, PicassoNet (TNNLS23) \cite{picassonet}} & 1.59 \\
      \hline
        \rowcolor{my_color}\makebox[5.0cm][l]{$\diamond$\, \textbf{Proto-Former (ours)}} & \textbf{1.47} \\
      \bottomrule[1pt]
  \end{tabular}
  \label{aflwtable}
  \vspace{-2.5em}
\end{table}


\textbf{Evaluation Metrics}: Normalized Mean Error (NME) is a widely used metric to evaluate the accuracy of face alignment. Specifically, the inter-pupil distance  $\rm NME_{ip}$ is used for COFW, the inter-ocular distance $\rm NME_{io}$ is applied for 300W and WFLW, and the bounding box size $\rm NME_{box}$ is used for AFLW. We also report the failure rate(FR)\cite{wu2018look} for COFW dataset.

\textbf{Compared Methods}: We compare our Proto-Former with several representative FLD methods including: LAB\cite{wu2018look}, AWing\cite{awing}, LUVLI\cite{kumar2020luvlifacealignmentestimating}, SAAT\cite{saatzhu2021improving}, ADNet\cite{huang2021adnet}, ODN\cite{odn}, LGSA\cite{tmm9082841}, PIPNET\cite{pipnetJLS21},
SLPT\cite{xia2022sparse}, GlomFace\cite{glomfacezhu2022occlusion}, DTLD\cite{DTLD}, ATF\cite{9749863_atf_tmm}, EfficentFan\cite{efficientfan}, PicassoNet\cite{picassonet}, Lite-HRNet\cite{Lite-HRNet}, Liang et al. \cite{liang2024}, MMDN\cite{mmdn}, DSLPT-R50\cite{10079153}, CIT-v2\cite{li2023cascaded}, HRNET\cite{hrnetSunXLW19}, DeCaF\cite{DeCaFA} and DAG\cite{dag}. For a fair comparison, the results are taken from the respective papers.

\textbf{Implementation Details}: In our experiments, the size of the input image is $512\times 512\times 3$. The weights $\lambda_1$, $\lambda_2$ and $\lambda_3$ are set to 1, 5, and 0.01, respectively. To improve the model’s robustness, we employ data augmentation techniques, including random image rotations of up to 30° and horizontal flips with a 50\% probability. Followed DETR, we utilize ResNet \cite{he2016deep} as the backbone network. The proposed Proto-Former is implemented in PyTorch and trained on an Nvidia RTX 4090 GPU for 100 epochs with a batch size of 8, using AdamW as the optimizer and an initial learning rate of $5 \times 10^{-5}$.

\subsection{Evaluations under Normal Circumstances}

Under normal conditions, we conduct comparative experiments on the Common Subset and Fullset of the 300W dataset, which mainly contain favorable facial images. Table \ref{300wtable} shows that our method achieves 2.61 $\rm NME_{io}$ on the 300W Common Subset and 2.95 $\rm NME_{io}$ on the 300W Fullset. Although Proto-Former is the coordinate regression FLD, it outperforms both SOTA heatmap regression FLD methods \cite{awing,wu2018look,kumar2020luvlifacealignmentestimating} and coordinate regression FLD methods \cite{xia2022sparse,glomfacezhu2022occlusion,odn}. Fig.\ref{exp_res1} visualizes the prototypes generated by APE. As seen in rows 1–4 and columns (D) and (E), clear facial contours appear in the 300W and WFLW datasets, but are less pronounced in COFW and AFLW. This indicates APE’s ability to capture dataset-specific structural features while suppressing irrelevant information.


\subsection{Evaluation of Robustness against Occlusion}

To evaluate the performance of our Proto-Former under occlusions, we conducted experiments on datasets such as COFW dataset, the 300W challenging Subset, and WFLW occlusion Subset. On the COFW test set (Table \ref{cofwtable}), Proto-Former achieves a  $\rm NME_{ip}$  of 4.66 and a failure rate of 0.2. On the 300W Challenging Subset, it achieves $\rm NME_{io}$ of 4.39 (Table \ref{300wtable}). Additionally, on the WFLW Occlusion Subset, it reaches a  $\rm NME_{io}$ of 5.00 (Table \ref{tabwflw}). The above experimental results demonstrate the effectiveness of the proposed Proto-Former under occluded scenarios. As shown in Fig. \ref{exp_res1}, the APE block adaptively selects prototype experts under occlusion, producing complementary high- and low-resolution prototypes. The former captures global structural context with a larger receptive field (Fig. \ref{exp_res1}(E)), while the latter preserves fine-grained local details (Fig. \ref{exp_res1}(C)). Their synergy interaction strengthens APAE’s facial geometry representation, so that Proto-Former can robustly capture key facial features even under severe occlusion.

\subsection{Evaluation of Robustness against Large Poses}
Facial images with large pose variations pose significant challenges for FLD. To assess the model’s performance under such conditions, we conducted experiments on the AFLW-Full test set, WFLW Pose Subset, and 300W Challenging Subset. Proto-Former achieves a $\rm NME_{io}$ of 4.39 on the 300W Challenging Subset (Table \ref{300wtable}) and 7.09 on the WFLW Pose Subset (Table \ref{tabwflw}), respectively, outperforming current state-of-the-art approaches \cite{wu2018look,wing,DeCaFA,mmdn,glomfacezhu2022occlusion,efficientfan,picassonet}. On the AFLW-Full test set, it attains an $\rm NME_{box}$ of 1.47, the second-best result, slightly inferior to DTLD \cite{DTLD}, mainly due to its two-stage architecture trained from scratch, compared to DTLD’s pretrained ResNet-18 with strong hierarchical priors. Fig.\ref{exp_res1} also display the corresponding prototypes. It can be seen that even under significant facial pose variations, the high-resolution prototype can effectively extract precise structural features from the high-resolution feature map by leveraging APE. This is likely because APE adaptively selects prototype experts that focus on profile regions, ensuring robust performance even under large pose deviations.

 \begin{table}[t]
  \centering
  \scriptsize
  \caption{Influence of APE and progressive prompt learning on the 300W Challenging Subset.}
  \renewcommand\arraystretch{0.4}
  \vspace{-1.3em}
  \begin{tabular}{c|cccc|c}
      \toprule[1pt]
      Method
      & TB
      & APE
      & PG 
      & ${\mathbb{L}_{PA}}$ 
      & \bf $\mathrm{NME_{io}}$  
      \\
      \midrule
      $\text{Trans (baseline)}$   
      & \ding{51}  &   &   &   & 4.66 \\

      $\text{Trans+APE}$
      & \ding{51} & \ding{51} &   &   & 4.60 \\

      $\text{Trans+APE+${\mathbb{L}_{PA}}$}$ 
      & \ding{51} & \ding{51} &   & \ding{51} & 4.54 \\

      $\text{Trans+APE+PG}$
      & \ding{51} & \ding{51} & \ding{51} &   & 4.42 \\

      \rowcolor{my_color}$\text{Trans+APE+PG+${\mathbb{L}_{PA}}$}$
      & \ding{51} & \ding{51} & \ding{51} & \ding{51} & 4.39 \\
      \bottomrule[1pt]
  \end{tabular}
  \label{tab_ape_ppb} 
  \vspace{-2em}
\end{table}
\subsection{Evaluation of Robustness against Blur}
This part focuses on facial images with varying blur, and experiments are conducted on the WFLW-full dataset and WFLW-blur Subset. On the WFLW-Full dataset, our Proto-Former achieves an $\rm NME_{io}$ of 4.23 on the test set, obtaining the second-best performance, slightly inferior to DAG \cite{dag}, as shown in Table \ref{tabwflw}. On the WFLW-Blur subset, Proto-Former attains an $\rm NME_{io}$ of 4.94, worse than DAG’s 4.82, mainly because DAG’s explicit graph reasoning better maintains spatial consistency, while Proto-Former’s implicit prototype-based learning is more susceptible to visual degradations (e.g. blur). As shown in Fig. \ref{exp_res1}, although the backbone feature maps (B) and (D) contain substantial irrelevant noise, Proto-Former effectively suppresses it via APE, yielding clearer and more defined prototypes in (C) and (E). By combining low- and high-resolution prototypes, the Proto-Encoders effectively capture coarse-to-fine structural representations. Through the collaboration of multiple prototype experts, clear prototypes can be extracted from noisy features, even in blurred facial images.

\subsection{Ablation Study}
The ablation studies will be conducted from the following aspects: influence of APE and prompt generator and influence of multi-datasets joint training. We show the details as follows.

\subsubsection{Influence of Adaptive Prototype Extractor and Prompt Generator}
The APE, prompt generator (PG) and $\mathbb{L}_{PA}$ are separately added to the baseline Trans (as shown in Table \ref{tab_ape_ppb}) for constructing our Trans+APE, Trans+APE+$\mathbb{L}_{PA}$, Trans+APE+PG and Trans+APE+PG+$\mathbb{L}_{PA}$. These models are tested on 300W challenging Subset respectively. From Table \ref{tab_ape_ppb}, we can see that $\text{Trans+APE+PG+$\mathbb{L}_{PA}$}$ surpasses $\text{Trans+APE+${\mathbb{L}_{PA}}$}$, $\text{Trans+APE+$\mathbb{L}_{PA}$}$ outperforms $\text{Trans}$ and $\text{Trans+APE+PG}$ exceeds $\text{Trans+APE+${\mathbb{L}_{PA}}$}$. These results can be attributed to: 1) The introduced APE significantly enhances the model’s adaptability to diverse facial structures by combining multi-scale prototypes (i.e., integrating both local and global prototypes) that used to generate refined prototypes. 2) The prompt generator further enhances the Proto-Decoder’s decoding process by producing highly relevant prompts from specific facial regions. 3) The PA loss function addresses the inconsistency in prototype expert activation during multi-dataset training, effectively alleviating gradient conflicts and ensuring stable learning. By integrating Trans, APE, PG, and $\mathbb{L}_{PA}$, the model achieves high-precision FLD across different datasets.



            

\subsubsection{Influence of Multi-datasets Training}
We conduct experiments using different dataset combinations. Starting with 300W as the baseline, we progressively add the AFLW, WFLW, and COFW datasets, achieving performance gains of 0.18, 0.44, and 0.57, respectively (as shown in Table \ref{tab7}). These results demonstrate the remarkable effectiveness of multi-datasets training.

\begin{figure}[t]
\begin{center}
	\includegraphics[width=0.88\linewidth]{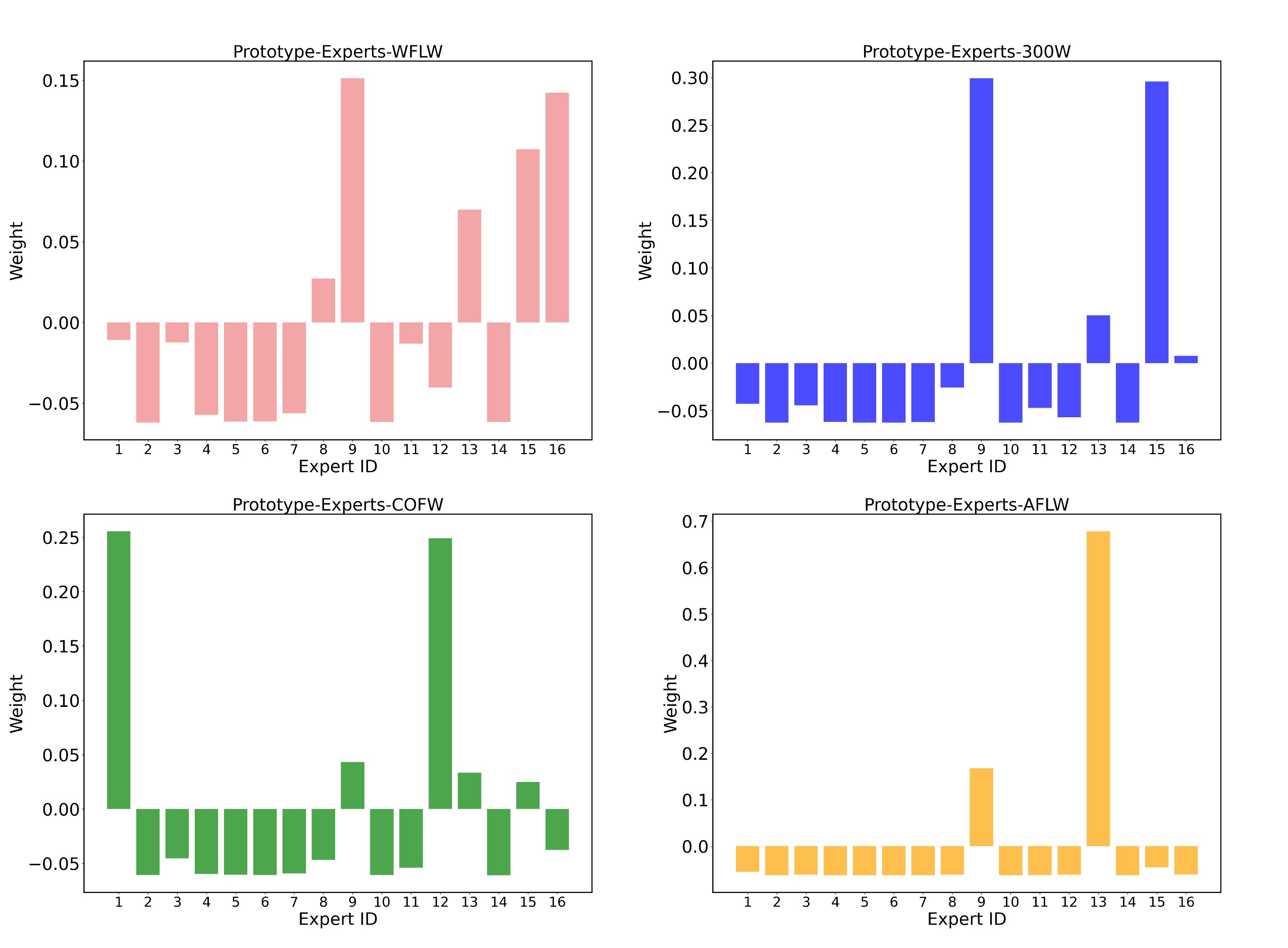}
	\end{center}
	\vspace{-1em}
	\caption{Visualization of the prototype experts selected by the APE and the normalized gating scores for different datasets.}
	\label{experts}
	\vspace{-1em}    
\end{figure}

\begin{figure}[t]
\begin{center}
	\includegraphics[width=0.88\linewidth]{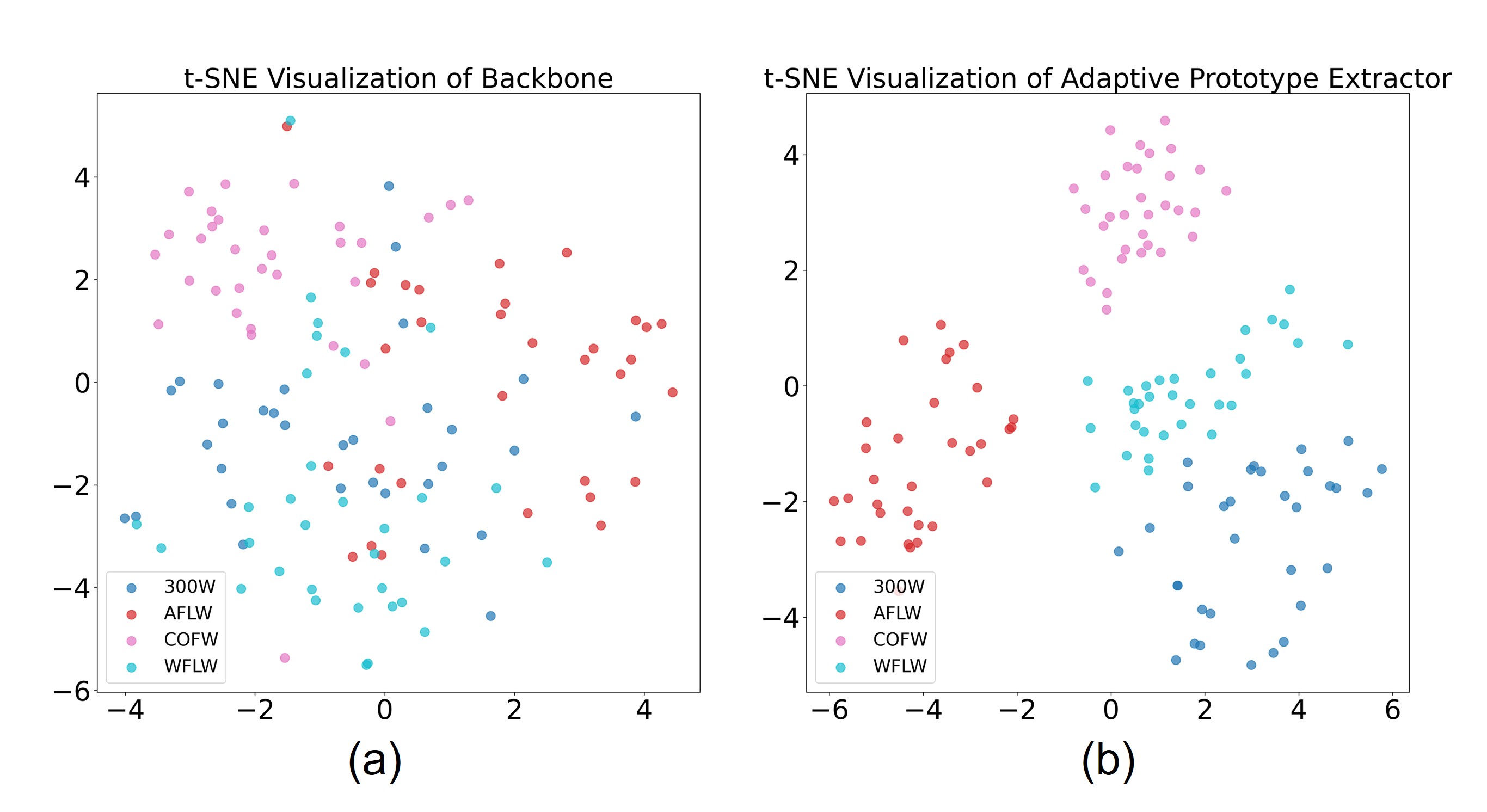}
	\end{center}
	\vspace{-1em}
	\caption{Comparison of t-SNE of backbone and APE. It shows that the APE can effectively distinguish the sample features of different datasets.}
	\vspace{-2em}    
\label{tsne}
\end{figure}

\begin{table}[t]
  \centering
  \scriptsize
  \caption{Influence of multi-dataset training on the 300W Challenging Subset.}
  \vspace{-1.4em}
  \renewcommand\arraystretch{0.4}
  \begin{tabular}{cccc|c}
      \toprule[1pt]
       \multicolumn{1}{c}{300W}
      & AFLW
      & WFLW
      & COFW
      & \bf $\mathrm{NME_{io}}$  
      \\
      \midrule
       \ding{51}  &   &   &   & 4.96 \\
       \ding{51} & \ding{51} &   &   & 4.78 \\
       \ding{51} & \ding{51} & \ding{51} &   & 4.52 \\
      \rowcolor{my_color}
       \ding{51} & \ding{51} & \ding{51} & \ding{51} & 4.39 \\
      \bottomrule[1pt]
  \end{tabular}
  \label{tab7}
  \vspace{-2em}
\end{table}
\begin{table}[t]
  \centering
  \scriptsize
  \caption{Influence of different numbers of experts and K values on the 300W challenging Subset (\% omitted).}
  \vspace{-1.5em}
  \renewcommand\arraystretch{0.4}
  \begin{tabular}{cc|cc}
      \toprule[1pt]
       \multicolumn{1}{c}{Number of experts}
      & K
      & $\mathrm{NME_{io}}$
      &  Params (M)
      \\
      \midrule
        18 & 8 & 4.42 & 62.70 \\
        14 & 8 & 4.46 & 60.71 \\
        10 & 8 & 4.46 & 58.73 \\
         6 & 3 & 4.51 & 56.74 \\
         4 & 2 & 4.52 & 55.75 \\
         2 & 1 & 4.53 & 54.76 \\
        \rowcolor{my_color}16 & 8 & 4.39 & 61.70 \\
      \bottomrule[1pt]
  \end{tabular}
  \label{expert_num}
  \vspace{-2em}
\end{table}

\subsection{Self Evaluation}
\subsubsection{Evaluation of different numbers of prototype experts and K values}
We investigated the effect of varying the number of activated experts. As shown in \ref{expert_num}, the model achieves its optimal performance with 16 experts at K=8. Reducing the number of experts from 16 to 2 leads to a gradual performance decline, likely due to insufficient diversity that constrains the model’s ability to capture complex data patterns. In contrast, increasing the number of experts from 16 to 18 also degrades performance, which may stem from redundancy or fragmented feature extraction that undermines the model’s learning capacity.

\subsubsection{Evaluation on Prototype Experts}
To illustrate the contribution of prototype experts to feature processing across different datasets, we visualize the expert paths for samples from each dataset, along with the corresponding weights of each prototype expert, based on the following formula:
\begin{equation}
   \tilde{\mathcal{G}}=\mathcal{G}-\frac{\sum_{i=1}^{N_\beta} \mathcal{G}_i}{N_\beta}
\end{equation}
where $\tilde{\mathcal{G}}$ represents the normalized weight of all prototype experts, $\mathcal{G}_i$ denotes the $i$-th sample's gating scores, $\mathcal{G}$ means the all samples' gating scores, ${N_\beta}$ denotes the number of dataset-specific landmarks. As shown in Fig. \ref{experts}, the diverse utilization of prototype experts indicates that facial structure reconstruction relies on distinct experts. Meanwhile, the variation in $\tilde{\mathcal{G}}$ across datasets demonstrates the experts' ability to dynamically adapt to different data distributions. This highlights their capacity to model facial prototypes based on the unique structural characteristics of each dataset.


\subsubsection{Evaluation on Adaptive Prototype Extractor}
In order to verify that the APE can adaptively process the facial structural features from different datasets samples (i.e., process these difficult-to-distinguish features into easily distinguishable facial structural features). we utilized t-SNE visualization to compare the performance of the Backbone and APE in processing these features. As shown in Fig.\ref{tsne} (a), the significantly overlap features from multiple datasets indicating that the sample feature distributions extracted by the Backbone are challenging to differentiate. In contrast, the clear clustering of multi-dataset samples in Fig.\ref{tsne} (b) demonstrates that the facial structural features represented by the prototypes after processing through the APE are easily distinguishable. These findings highlight the APE’s ability to construct a dynamic routing space for the adaptive processing of facial structural features.


\begin{table}[t]
  \centering
  \scriptsize
  \caption{Comparison of computational complexity and inference efficiency.}
  \vspace{-1.4em}
  \renewcommand\arraystretch{0.55}
  \begin{tabular}{c|ccc}
      \toprule[1pt]
       Method
        & Params (M) & FLOPs(G) & FPS (frames/s)
      \\
      \midrule
      $\text{Trans (baseline)}$ & 39.99 & 34.54 & 30.87 \\
        $\text{Trans+APE}$        & 60.04 & 42.56 & 24.04 \\
        \rowcolor{my_color} $\text{Trans+APE+PG}$ & 61.70 & 44.07 & 22.05 \\
      \bottomrule[1pt]
  \end{tabular}
  \label{tab_time_ana}
  \vspace{-2em}
\end{table}


\subsubsection{Time and memory analysis} 
Inspired by DETR \cite{carion2020end}, the proposed Proto-Former introduces a multi-level encoder equipped with an Adaptive Prototype Extractor (APE) to establish the APAE, and integrates PG to develop a progressive prompt learning–based decoder (PPAD). As reported in Table \ref{tab_time_ana}, Proto-Former incurs higher parameter and computational overhead than the baseline. The baseline model (Trans) contains 39.99M parameters, which increase to 60.04M with APE (Trans+APE) and further to 61.70M with the addition of PG (Trans+APE+PG). On a single RTX 3060 12GB GPU, Proto-Former achieves an inference speed of 22.05 FPS, which increases to 24.04 FPS when the PG is removed. In terms of computational complexity, Proto-Former requires 44.07 GFLOPs, whereas the variant without the PG requires 42.56 GFLOPs. While Proto-Former incurs additional parameters and computational overhead, the improvements in performance are considerable. Moreover, these costs are expected to become negligible with future hardware and software advancements.

\begin{table}[t]
  \centering
  \scriptsize
  \caption{The effect of different $\lambda$ settings on the 300W Challenging Subset (\% omitted).}
  \vspace{-1.5em}
  \renewcommand\arraystretch{0.4}
  \begin{tabular}{ccc|c}
      \toprule[1pt]
       \multicolumn{1}{c}{$\lambda_1$}
       & $\lambda_2$ & $\lambda_3$ & $\mathrm{NME_{io}}$
      \\
      \midrule
        1 & 5 & 10    & 4.60 \\
        1 & 5 & 1     & 4.54 \\
        1 & 5 & 0     & 4.42 \\
        1 & 5 & 0.1   & 4.47 \\
        1 & 5 & 0.05  & 4.47 \\
        1 & 5 & 0.001 & 4.42 \\
        \rowcolor{my_color} 1 & 5 & 0.01 & 4.39 \\
      \bottomrule[1pt]
  \end{tabular}
  \label{tab_sensi}
  \vspace{-2em}
\end{table}


\subsubsection{Sensitivity analysis of parameters}  The overall training loss is composed of the $\mathbb{L}_{coor}$, $\mathbb{L}_{index}$, and $\mathbb{L}_{PA}$. Following \cite{carion2020end}, the weighting coefficients $\lambda_1$ and $\lambda_2$ are set to 1 and 5, respectively. As shown in Table \ref{tab_sensi}, we report the Proto-Former’s $\rm NME_{io}$ on the 300W Challenging Subset under different settings of the weighting parameter $\lambda_3$. The results indicate that the model achieves the best performance with $\lambda_3$=0.01, attaining an $\rm NME_{io}$ of 4.39. For other values of $\lambda_3$, i.e., 0.001, 0.05, 0.1, 0, 1, and 10, the corresponding $\rm NME_{io}$ are 4.42, 4.47, 4.47, 4.42, 4.54 and 4.60. Hence, $\lambda_3$ is selected as the optimal weighting strategy for model training.


\section{Conclusion}
In the UFLD task, leveraging a unified model to extract dataset-specific features remains a challenging problem. This paper proposes Proto-Former, which employs a multi-dataset training strategy and seamlessly integrates APAE, PPAD and ${\mathbb{L}_{PA}}$ to address the above challenge. Experimental results show that the APAE not only establishes a dynamic routing space and extracts prototypes through the APE but also uses Proto-Encoders to effectively refine prototype features. thereby enhancing the decoding efficiency of the prototype decoder and achieving high-precision FLD. The PA loss imposes constraints on the activation distribution of prototype experts, effectively preventing overly dispersed activations. This reduces interference among characteristics of different datasets, ultimately alleviating  the issue of gradient conflicts. Experiments on four popular FLD datasets demonstrate that our proposed Proto-Former outperforms the current SOTA methods.

\bibliographystyle{IEEEtran}
\bibliography{IEEEabrv,IEEEexample}
\vspace{-6em}   
\begin{IEEEbiography}
[{\includegraphics[width=1in,height=1.25in,clip,keepaspectratio]{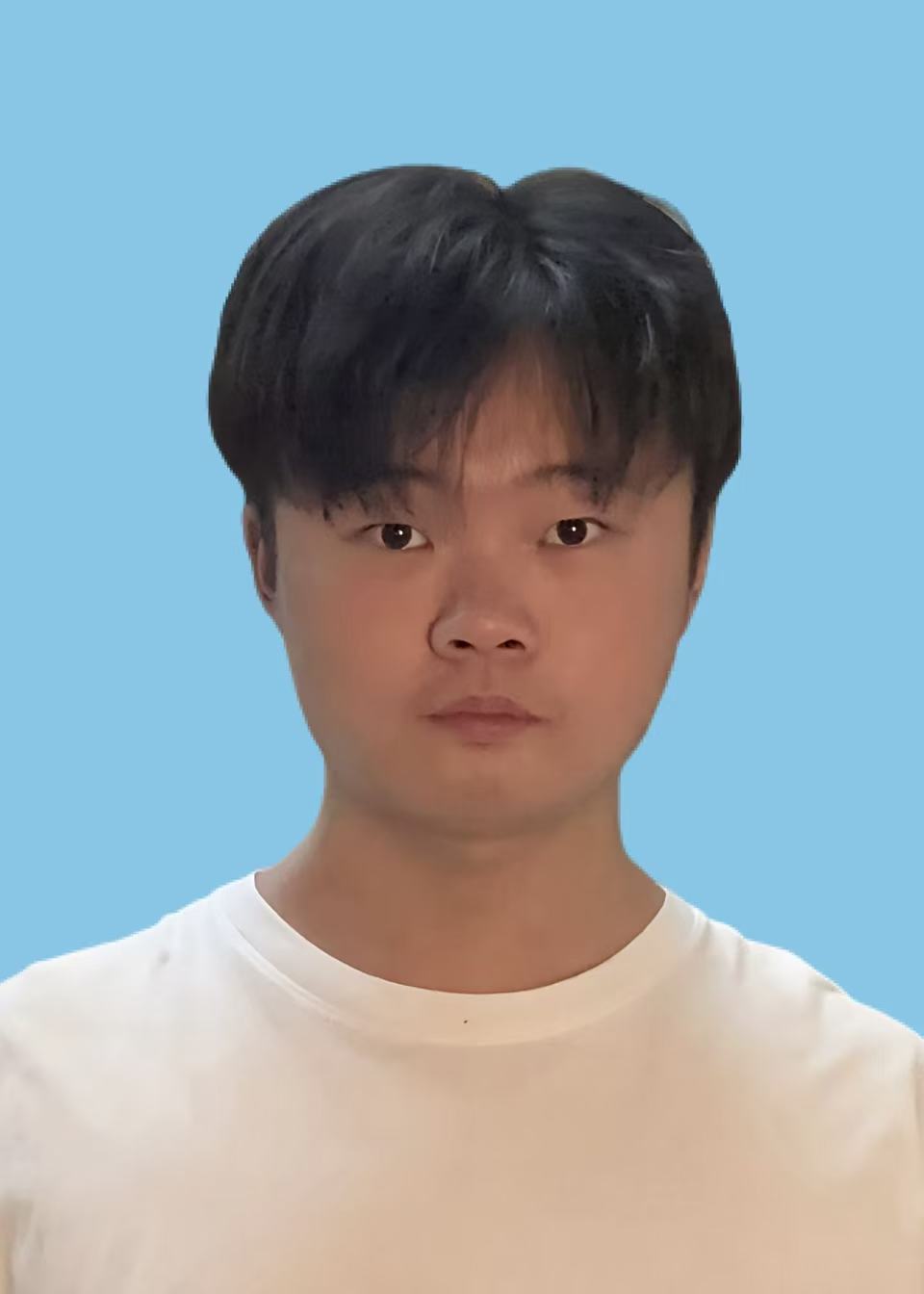}}]{Shengkai Hu}
is currently pursuing the M.S. degree with Zhongnan University of Economics and Law, Hubei, China. His research interests include facial landmark detection and image restoration.
\end{IEEEbiography}
\vspace{-6em}   

\begin{IEEEbiography}[{\includegraphics[width=1in,height=1.25in,clip,keepaspectratio]{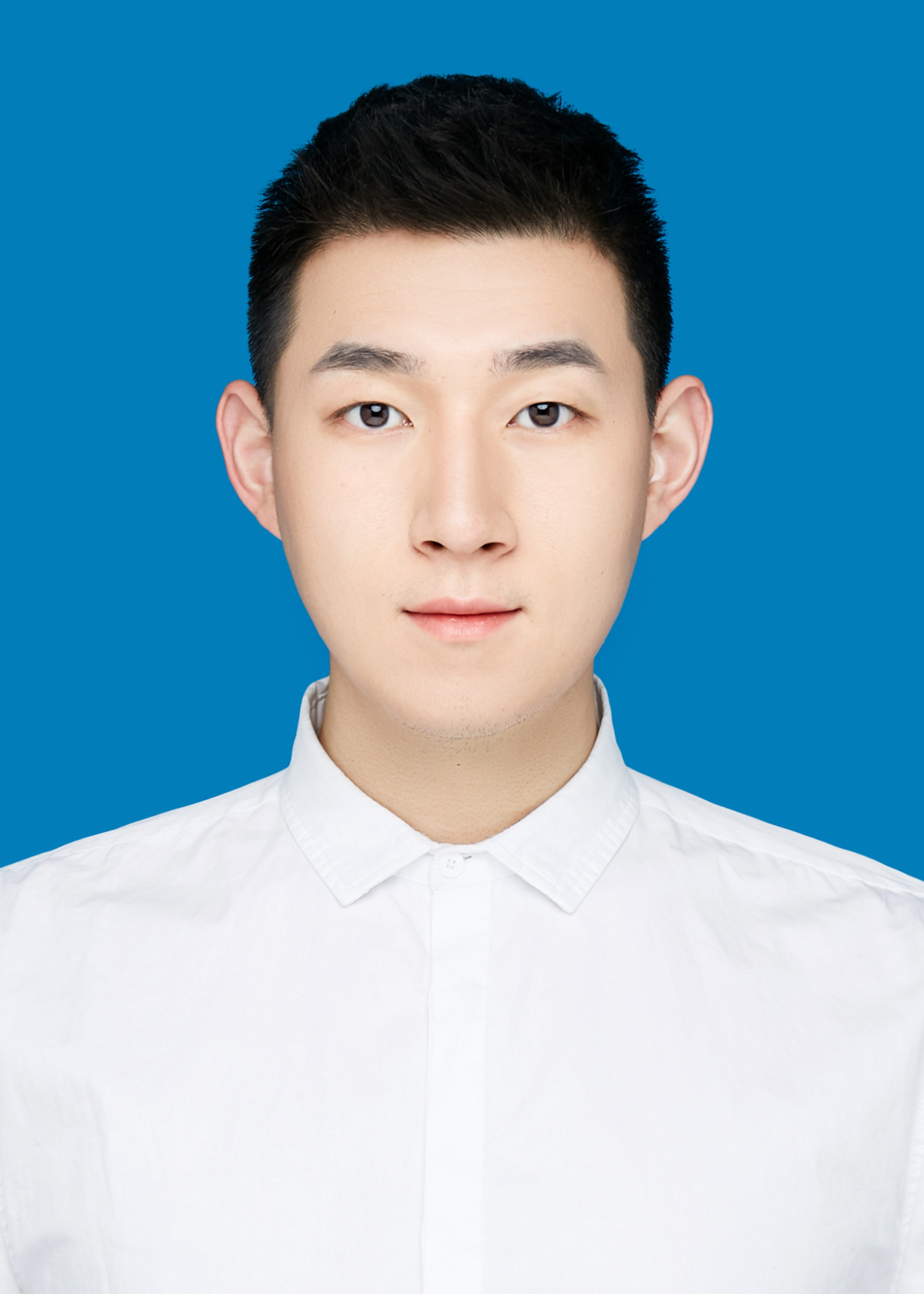}}]{Haozhe Qi}
is currently pursuing the M.S. degree with Zhongnan University of Economics and Law, Hubei, China. His research interests include image processing and computer vision.
\end{IEEEbiography}
\vspace{-4em}   

\begin{IEEEbiography}
[{\includegraphics[width=1in,height=1.25in,clip,keepaspectratio]{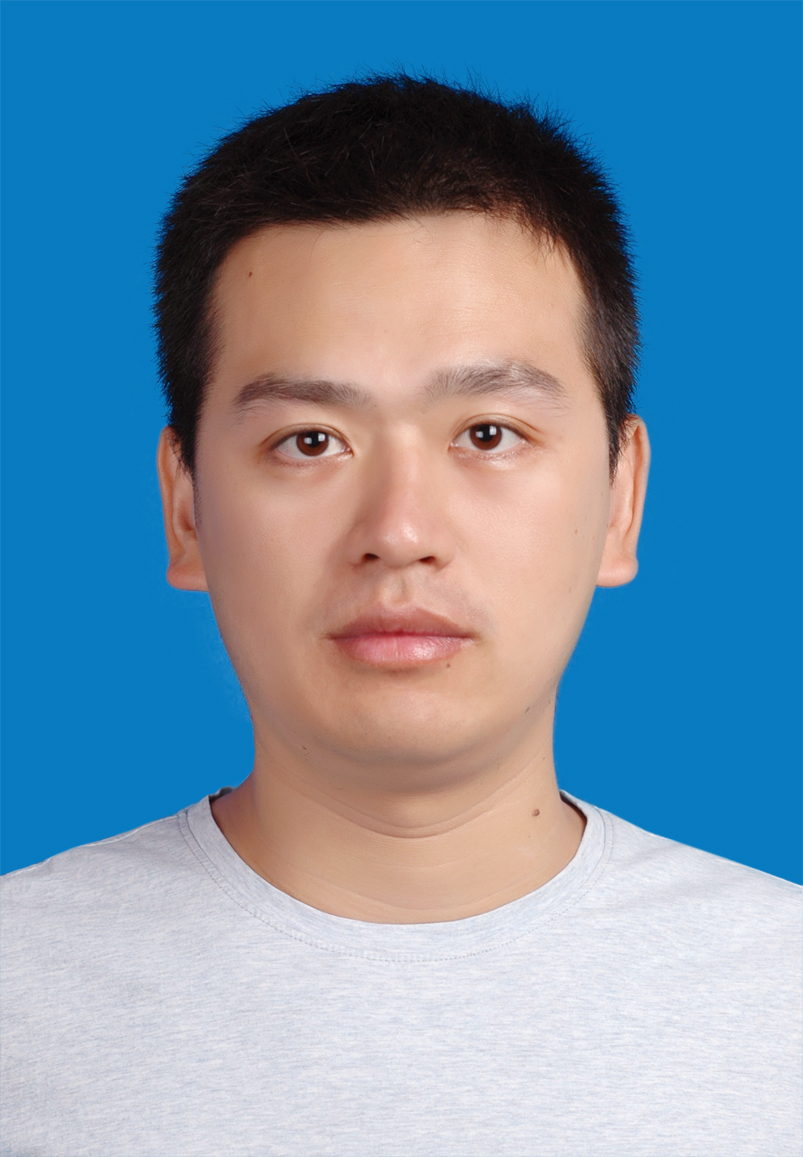}}]{Jun Wan}
	received the Ph.D. degree in School of Computer Science, Wuhan University, China, in 2019. From 2019 to 2021, He was a Post-Doctoral Fellow with the College of Computer Science and Software Engineering, Shenzhen University, China. He is now an Associate Professor in the School of Information Engineering, Zhongnan University of Economics and Law, Wuhan, 430073, China, and also a Visiting Scholar with the College of Computing and Data Science, Nanyang Technological University, Singapore. His main research interests include computer vision, landmark detection and image/video captioning. His works have been published in premier computer vision journals and conferences, including IJCAI, TIP, TCYB, TKDE, TNNLS, TFS, Neural Networks, Pattern Recognition, Information Sciences and so on.
\end{IEEEbiography}
\vspace{-4em}   

\begin{IEEEbiography}[{\includegraphics[width=1in,height=1.25in,clip,keepaspectratio]{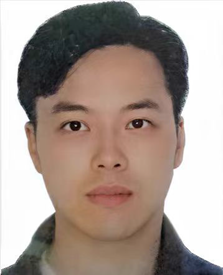}}]{Jiaxing Huang} (Member, IEEE) received his B.Eng. in EEE from University of Glasgow, UK, and PhD from Nanyang Technological University (NTU), Singapore. He is currently a Research Fellow with College of Computing and Data Science, NTU. His research  include computer vision and machine learning.
\end{IEEEbiography}
\vspace{-4em}   

\begin{IEEEbiography}[{\includegraphics[width=1in,height=1.25in,clip,keepaspectratio]{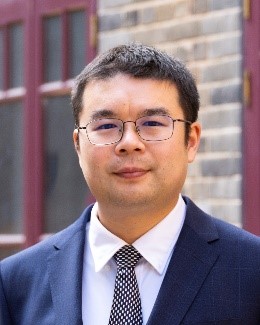}}]{Lefei Zhang}received the B.S. and Ph.D. degrees from Wuhan University, Wuhan, China, in 2008 and 2013, respectively. He was a Big Data Institute Visitor with the Department of Statistical Science, University College London, U.K., and a Hong Kong Scholar with the Department of Computing, The Hong Kong Polytechnic University, Hong Kong, China. He is a professor with the School of Computer Science, Wuhan University, Wuhan, China, and also with the Hubei Luojia Laboratory, Wuhan, China. His research interests include pattern recognition, image processing, and remote sensing. Dr. Zhang serves as a topical editor of IEEE Transactions on Geoscience and Remote Sensing, an associate editor of Pattern Recognition, and a section editor-in-chief of Remote Sensing.
\end{IEEEbiography}
\vspace{-4em}   

\begin{IEEEbiography}[{\includegraphics[width=1in,height=1.25in,clip,keepaspectratio]{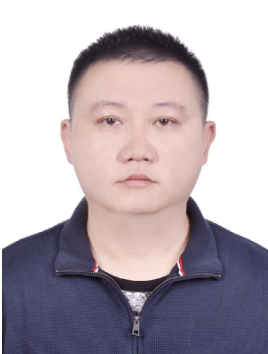}}]{Hang Sun}received the Ph.D. degree from the School of Computer Science, Wuhan University, Wuhan, China, in 2017. He is currently an Associate Professor with the College of Computer and Information Technology, China Three Gorges University, Yichang, China. His research include computer vision and image restoration.

\end{IEEEbiography}
\vspace{-4em}   

\begin{IEEEbiography}[{\includegraphics[width=1in,height=1.25in,clip,keepaspectratio]{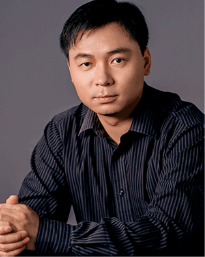}}]{Dacheng Tao}(Fellow, IEEE) is currently a Distinguished University Professor in the College of Computing \& Data Science at Nanyang Technological University. He mainly applies statistics and mathematics to artificial intelligence and data science, and his research is detailed in one monograph and over 200 publications in prestigious journals and proceedings at leading conferences, with best paper awards, best student paper awards, and test-of-time awards. His publications have been cited over 112K times and he has an h-index 160+ in Google Scholar. He received the 2015 and 2020 Australian Eureka Prize, the 2018 IEEE ICDM Research Contributions Award, and the 2021 IEEE Computer Society McCluskey Technical Achievement Award. He is a Fellow of the Australian Academy of Science, AAAS, ACM and IEEE.
\end{IEEEbiography}

\end{document}